\documentclass[11pt]{article}

\usepackage[final]{acl}

\usepackage{times}
\usepackage{latexsym}

\usepackage[T1]{fontenc}

\usepackage[utf8]{inputenc}

\usepackage{microtype}

\usepackage{inconsolata}

\usepackage{graphicx}
\usepackage[linesnumbered,ruled,vlined]{algorithm2e}
\usepackage{amsmath} 
\usepackage{booktabs} 

\usepackage{ltablex}  
\usepackage{caption}
\usepackage{geometry}

\usepackage{longtable}
\usepackage{siunitx}

\usepackage{amsfonts}
\usepackage{multirow}
\usepackage{float}

\title{Polarity Classification with Multitask Learning Syntax Augmentation}

\author{
 \textbf{Muhammad Imran\textsuperscript{1,*}},
 \textbf{Carlos Gómez-Rodríguez\textsuperscript{1}}
\\
\\
 \textsuperscript{1}  Universidade da Coruña, CITIC, \\
  Depto. de Ciencias de la Computación y Tecnologías de la Información, \\
  Campus de Elviña s/n, 15071 A Coruña, Spain \\
\\
 \small{
   \textsuperscript{*} \textbf{Corresponding author:} \href{mailto:m.imran@udc.es}{m.imran@udc.es}
 }
}

\begin{document}
\maketitle
\begin{abstract}
Sentiment Analysis (SA) is a fundamental NLP task concerned with the identification and categorization of subjective information in text. While lexicon-based methods offer transparency, they often struggle with linguistic phenomena such as negation and intensification. Syntactic parsing has been integrated into SA pipelines to address this, but conventional parsers introduce significant computational overhead. Recent work shows that reformulating dependency parsing as a Sequence Labeling (SEL) task can improve efficiency. We propose and compare two strategies for integrating syntactic knowledge into a neural sentiment analysis model under unified Multi-Task Learning (MTL). The first, Standard-MTL, jointly learns dependency parsing (via sequence labeling) and sentiment polarity classification using a shared encoder with two task-specific heads. The second, Contrastive-MTL, provides implicit syntactic guidance through a contrastive auxiliary objective and a syntax-biased attention pooling mechanism, without requiring explicit label prediction. Both rely on syntactic annotations during training and eliminate need for hand-crafted compositional rules or external parsing at inference. We evaluate eleven datasets spanning seven languages---the MARC corpus (English, Spanish, French, German) and seven datasets from SemEval-2022 Task 10 (English, Spanish, Norwegian, Basque, Catalan)---using four established SEL encodings. With the relative encoding, Contrastive-MTL outperforms a syntax-naive baseline on 10 of 11 datasets and surpasses the external rule-based pipeline (SELSP) that parses at inference on all six datasets where the latter applies, while being 1.5--9$\times$ faster at evaluation. Standard-MTL is more dataset- and encoding-dependent. These results show implicit syntactic supervision during training can match or exceed explicit parsing at inference without computational overhead.
\end{abstract}

\section{Introduction}

Sentiment Analysis (SA) is a fundamental task in Natural Language Processing (NLP) that deals with the identification and categorization of subjective information in a given text, typically involving subjectivity detection and polarity classification (e.g., positive, negative, or neutral) \cite{cui2023survey}. 
In this article, we focus on sentiment polarity classification: given an input text \(x\) (a sentence or document), predict its overall sentiment polarity \(y \in \mathcal{Y}\), where \(\mathcal{Y}\) typically contains labels such as positive, negative, and neutral. This is a sentence- or document-level classification task. We do not address fine-grained tasks like structured or aspect-based sentiment analysis.

Lexicon-based approaches leverage sentiment dictionaries like SO-CAL \cite{brooke-etal-2009-cross} to detect polarity-bearing words, providing an intuitive and transparent approach to SA. The explainability of lexicon-based methods stands in contrast to that of ``black-box'' supervised machine learning models. In lexicon-based approaches, the predicted sentiment can be traced directly to specific lexical items, whereas supervised models, despite often achieving high accuracy, obscure the reasoning behind their predictions \cite{cui2023survey}. 

However, the accuracy of lexicon-based approaches depends on the handling of linguistic phenomena such as negation (``not good'') and intensification (``very good''), which have the tendency to shift or modify sentiment polarity. The scope of these linguistic phenomena is quite complex. To handle it properly, researchers have integrated syntactic parsing within rule-based SA systems \cite{Vilares2017, imran2026syntax}. By analyzing the syntactic structure of a sentence-often represented via Universal Dependencies (UD) \cite{Zeman2023}, rule-based systems can apply recursive, compositional rules to accurately determine the scope of polarity-shifting elements. Despite these advantages in accuracy and explainability, conventional syntactic parsers, such as Stanza \cite{qi2020stanza}, are computationally expensive, creating a bottleneck for large-scale or real-time SA applications \cite{gomez2018constituent, strzyz2019viable}.

A promising solution to this computational bottleneck is the reformulation of dependency parsing as a Sequence Labeling (SEL) task \cite{gomez2018constituent, strzyz2019viable}. SEL parsers rely on 
lightweight scripts
\cite{roca2022system} to linearize the tree structure into a sequence of labels, enabling the use of fast, off-the-shelf sequence taggers for syntactic analysis. Recent work by~\citet{imran2026syntax} successfully applied a Sequence Labeling Syntactic Parser (SELSP) within a rule-based sentiment analysis pipeline, demonstrating significant improvements in processing speed over traditional parsers while maintaining competitive accuracy for polarity classification.

While effective, the pipeline approach in \cite{imran2026syntax} treats parsing and sentiment analysis as two discrete stages: first, a full parse tree is generated via sequence labeling, and then a separate rule-based engine computes sentiment from that structure. In this article, we propose two alternative strategies for integrating syntactic knowledge under a unified \textbf{Multi‑Task Learning (MTL) framework}. Unlike prior work that uses a pipeline architecture---where a sequence labeling parser first generates a full parse tree, followed by a separate rule-based engine that computes sentiment from that structure---our proposed MTL models leverage syntactic annotations (generated by an accurate off-the-shelf parser) during training to provide an auxiliary learning signal, thereby eliminating the need for hand‑crafted compositional rules or external parsing at inference time. Our first variant, \textbf{Standard-MTL}, employs explicit multi‑task supervision by jointly training a shared encoder to predict sentiment polarity and dependency relation labels simultaneously. In contrast, our second variant, \textbf{Contrastive-MTL}, integrates syntax implicitly through a contrastive auxiliary objective and a syntax‑biased attention pooling mechanism. Crucially, both models share an identical baseline sentiment architecture; the auxiliary syntactic signals are discarded at inference to ensure that any performance gains stem solely from the internalized linguistic structure learned during training.

Recent advances in contrastive learning have demonstrated the value of both sentence-level and token-level objectives. Sentence-level methods such as SimCSE \cite{gao-etal-2021-simcse}, DeCLUTR \cite{giorgi-etal-2021-declutr}, and supervised contrastive learning \cite{khosla2020supervised} learn robust representations by pulling together semantically similar sentences and pushing apart dissimilar ones. More recently, token-level objectives have proven effective in different settings: MEXMA \cite{janeiro-etal-2025-mexma} combines sentence-level and token-level objectives for cross-lingual sentence encoders by predicting masked tokens across languages, while Critical Tokens Matter \cite{lin2024critical} uses token-level contrastive estimation to identify and penalize critical tokens in LLM reasoning trajectories. These works establish that token-level signals can shape representations more effectively than sentence-level objectives alone. However, these methods either operate at the sentence/instance level or target cross-lingual encoding and LLM reasoning; none uses dependency relations as token-level contrastive signals for sentiment analysis. Our Contrastive-MTL fills this gap by combining token-level syntax and sentiment contrastive losses with syntax-biased attention pooling, injecting syntactic knowledge without requiring parsing at the inference stage.

Our contributions are threefold: (1) We present two complementary multi‑task learning strategies for integrating syntactic information into sentiment analysis: Standard-MTL, which jointly learns dependency parsing as sequence labeling and sentiment classification, and MTL‑Contrastive, which injects syntactic guidance implicitly through a contrastive auxiliary objective and syntax‑biased attention pooling. 
(2) We evaluate both MTL variants on eleven datasets: MARC product reviews (spanning four language subsets) and seven SemEval-2022 Task~10 corpora covering hotel, news, and multi-domain reviews---across seven languages in total and using four sequence-labeling encodings. We then compare their performance against a single-task baseline (sentiment-only, without syntactic supervision), an external rule-based pipeline that parses at inference
(SELSP), and two ablations that isolate the contributions of the
contrastive losses and the attention bias.
(3) Through a mechanism analysis on a SemEval-2022 English dataset combining linear probing, inference-time attention analysis, and learned-bias inspection, we show that the contrastive objective produces encoder representations that are linearly separable by dependency relation, and
that the attention bias alone --- without the contrastive anchor --- degrades encoder-side syntactic encoding. 

\section{Related Work}
Methods for SA are broadly categorized into Machine Learning/Deep Learning (ML/DL) approaches, lexicon-based methods, and hybrid techniques.

\subsection{Supervised ML/DL methods}
 Supervised ML/DL methods, including models based on LSTMs, CNNs, and pre-trained Transformers like BERT and RoBERTa, represent the current state of the art in academic benchmarks \cite{cui2023survey}. These models excel when training and test data share similar characteristics but often require large, domain-specific annotated corpora and lack interpretability. Lexicon-based methods, conversely, utilize sentiment dictionaries and are prized for their simplicity, cross-domain robustness, and explainability, making them prevalent in industrial applications \cite{taboada2011lexicon, hutto2014vader, denecke2023sentiment}. Hybrid approaches seek to combine the strengths of both, for example, by using lexicon-derived features to augment the input to a neural classifier \cite{fu2018lexicon}.

\subsection{Syntactic Parsing for Sentiment Analysis}
To address the challenge of determining the scope of polarity-shifting elements like negation, a well-established line of research employs syntactic rules that operate on parse trees. Universal Dependencies (UD) \cite{Zeman2023} has emerged as the dominant framework for this purpose due to its cross-lingual consistency and extensive treebank resources. Systems like those described by \cite{Vilares2017} and \cite{imran2026syntax, kellert2025evaluating} use UD parse trees to apply bottom-up, compositional rules that modify and aggregate sentiment scores based on syntactic relations (e.g., adverbial modifiers for intensification, negation cues). This approach provides a linguistically motivated, transparent alternative to heuristic window-based scope detection \cite{Thelwall2010} or opaque ML classifiers for scope resolution \cite{singh2021deep}.

\subsection{Dependency Parsing as Sequence Labeling}
The computational overhead of traditional transition-based or graph-based dependency parsers has motivated the development of faster alternatives. By encoding a dependency tree into a sequence of labels that can be predicted by a simple tagger, parsing is reduced to a sequence labeling task \cite{gomez2018constituent}. The CoDeLin framework \cite{strzyz2019viable, roca2022system} supports various encoding schemes, including absolute, relative and POS-based offsets and a range of bracketing-based encodings, which
can be used to represent a dependency tree as a sequence of one label per word.
\cite{imran2026syntax} demonstrated that a SEL parser (SELSP) trained with these encodings could be integrated into a rule-based SA pipeline, substantially improving parsing speed with minimal loss in downstream sentiment accuracy compared to Stanza. 

\subsection{Multi-Task Learning in NLP}
Multi-Task Learning (MTL) is a paradigm in which a single model is trained to perform multiple related tasks simultaneously, leveraging shared representations to improve generalization and efficiency for all tasks \cite{van-der-goot-etal-2021-massive}. In NLP, MTL has been successfully applied to learn joint representations for a variety of tasks. \citet{Collobert2008} pioneered a unified architecture for tasks like part-of-speech tagging, chunking, and named entity recognition. More recently, MTL has been used to infuse syntactic knowledge into semantic tasks. For instance, \cite{swayamdipta2018syntactic} proposed a model that jointly learns syntactic chunking and semantic composition for sentence representations. Similarly, \cite{zhang2016stack} showed that training a parser jointly with a sentence-level classification task (like sentiment) can lead to improved parsing accuracy. In the domain of syntactic parsing itself, MTL has been employed to jointly predict part-of-speech tags, morphological features, and dependency trees \cite{kondratyuk201975}.

The incorporation of syntactic knowledge, ranging from dependency relations to constituent structures, has proven beneficial for handling complex linguistic phenomena in sentiment analysis, such as negation and intensification \cite{poria-etal-2014-sentic, gomez2024dancing}. 
Moreover, the effectiveness of parsing as sequence labeling has been further demonstrated in challenging multilingual settings, including code-switched and low-resource languages \cite{kellert-etal-2025-parsing}. In structured sentiment analysis, \citet{imran2026structured} linearize sentiment dependency graphs over holders, targets, and expressions and decode them through sequence labeling. However, that work targets fine-grained structured sentiment analysis: the sequence labeler predicts the graph structure itself, and the output at inference is a sentiment graph. In contrast, this study focuses on sentence/document-level sentiment polarity classification. Here, we use dependency parsing, formulated as sequence labeling, as an auxiliary training signal; the parser is discarded at inference, and the final output is a single polarity label for the input text. Thus, while both works exploit sequence labeling for dependency structures, they differ in task formulation, output space, and inference-time objective.

Beyond sentiment analysis, syntactic supervision via MTL has also proven beneficial for other sequence labeling tasks, such as biomedical named entity recognition \cite{imran2026synner}. Our work builds upon these principles by applying MTL specifically to the combination of dependency parsing (as sequence labeling) and sentiment polarity classification, offering a unified alternative to the two-stage rule-based pipelines such as \citet{imran2026syntax, kellert2025evaluating} which require parsing at inference time.

\subsection{Token-Level Contrastive Learning}
Contrastive learning in NLP has largely focused on sentence-level objectives. SimCSE \cite{gao-etal-2021-simcse} uses dropout as noise to create positive pairs, DeCLUTR \cite{giorgi-etal-2021-declutr} constructs positive pairs from different spans of the same document, and supervised contrastive learning \cite{khosla2020supervised} uses class labels to define positive pairs. Although, these methods learn powerful sentence representations but do not explicitly model token-level syntactic structures. 

Token-level objectives have recently shown promise: MEXMA \cite{janeiro-etal-2025-mexma} combined sentence-level and token-level objectives for cross-lingual sentence encoders by using the sentence representation in one language to predict masked tokens in another, improving bitext mining and downstream tasks. Critical Tokens Matter \cite{lin2024critical} introduced token-level contrastive estimation to identify ``critical tokens'' in LLM reasoning trajectories and penalized them via contrastive direct preference optimization, improving reasoning accuracy. Neither uses dependency relation labels as a contrastive signal nor syntax-guided attention for sentiment analysis. Our work differs by defining token-level contrastive losses conditioned on dependency relations and sentiment labels, combined with syntax-biased attention pooling. To our knowledge, this is the first such formulation for sentiment analysis.

\section{Data and Preprocessing}
 We performed experiments on the Multilingual Amazon Reviews Corpus (MARC)\footnote{\url{https://huggingface.co/datasets/mteb/amazon_reviews_multi/tree/main}} \cite{keung2020multilingual} a large collection of Amazon reviews for multilingual text in four languages (English, Spanish, French and German). Each review in this dataset provides a 5-star rating (1–5, mapped to labels 0–4 corresponding to very negative, negative, neutral, positive, and very positive, respectively) at the review level, making it well-suited for multilingual text classification. Furthermore, we performed experiments on the five datasets released in the SemEval-2022 Task 10: 
the OpeNER dataset \cite[OpN;][]{agerri-etal-2013-opener} with English and Spanish annotations, 
the English Multi-Perspective Question Answer corpus \cite[MPQA;][]{wiebe2005annotating} and the Darmstadt Universities corpus \cite[DS;][]{toprak-etal-2010-sentence}, the Norwegian NoReC dataset \cite{velldal-etal-2018-norec} and the Basque and Catalan MultiBooked dataset \cite[MB;][]{barnes-etal-2018-multibooked}.

In SemEval-2022 Task 10 datasets, every review is annotated with the individual polarity expressions and their polarity (positive or negative) as demonstrated by few simple examples of reviews in Table ~\ref{tab:examples} from \cite{barnes-etal-2022-semeval} that only contain one polarity expression. Since our problem of interest is polarity classification, we extract review-level  sentiment labels from the structured opinion annotations as described in  Algorithm~\ref{alg:convert}. Each review  in the dataset contains an \texttt{opinions} field consisting of tuples  $(P, I, E)$, where $P \in \{\text{Positive}, \text{Negative}, \text{Neutral}\}$  is the polarity, $I \in \{\text{Strong}, \text{Standard or Average}, \text{Weak}\}$, is the intensity, and $E$ is the polar expression span.  We discard reviews that contain no polarity expressions at all (i.e., the  \texttt{opinions} field is empty), so our evaluation is comprised only on reviews  that do contain subjectivity. For reviews with multiple polarity expressions, we  compute an aggregate review-level polarity label by scoring each opinion based on  both its polarity and intensity: Positive opinions contribute $+2$ if Strong and  $+1$ otherwise, Negative opinions contribute $-2$ if Strong and $-1$ otherwise,  and Neutral opinions are tracked but do not affect the score. The final gold  polarity label is determined by comparing the aggregated positive and negative  scores: the label is \textit{Positive} if the positive score exceeds the negative  score, \textit{Negative} if the negative score exceeds the positive score, and  \textit{Neutral} if the scores are tied and at least one non-neutral or neutral  opinion exists. We thus have a task of predicting three polarity labels (\textit{Positive}, \textit{Negative}, and \textit{Neutral}) on these datasets. The statistics of these datasets, including the number of discarded  and remaining reviews per split, are presented in Table~\ref{tab:dataset_stats}.


Each sentence is annotated with a review-level sentiment label. We used Biaffine Dependency Parser \cite{dozat2017deep} (dep-biaffine-xlmr) available in the SuPar package \footnote{\url{https://github.com/yzhangcs/parser}} to generate dependency parse trees (dependency head \& relations). To cast dependency parsing as sequence labeling we relied on CoDeLin framework \cite{strzyz2019viable, roca2022system} which supports various encoding schemes
 to represent dependency trees as sequences of labels.
Following CoDeLin, each word $w_i$ in a sentence of length $n$ is assigned a label $(x_i, l_i)$, where $l_i$ is the dependency relation and $x_i$ encodes the position of its head. A dummy root node $0$ is used. We consider four encodings:
\begin{itemize}
    \item \textbf{Absolute}: $x_i$ is the absolute index of the head, as in the CoNLL format.
    \item \textbf{Relative}: $x_i$ is the head index minus the dependent index.
    \item \textbf{BrkBased}: $x_i$ is a string of bracketing symbols that encodes the tree structure. Arcs are represented by symbols such as \texttt{<}, \texttt{>}, \texttt{\textbackslash}, and \texttt{/} at fencepost positions between words, allowing the tree to be reconstructed with a stack.
    \item \textbf{Brk7Bits}: a compact variant of BrkBased in which each bracketing component is encoded as a fixed 7-bit integer, reducing the label space~\citep{gomez-rodriguez-etal-2023-4}.
\end{itemize}






\begin{table}[tbp]
\centering
\caption{\label{tab:examples} Polarity expression and corresponding polarity from the English training dataset in \citep{agerri-etal-2013-opener}}
\small
\begin{tabular}{p{0.55\columnwidth} c}
\toprule
\textbf{Review} & \textbf{Polarity} \\
\midrule
Clean , renewed rooms. & Positive \\
Food very basic. & Negative \\
What can we say. & No Annotation \\
\bottomrule
\end{tabular}
\end{table}

\begin{algorithm*}
\caption{Conversion of structured sentiment annotations to review-level polarity}
\label{alg:convert}
\SetAlgoLined
\KwIn{A dataset $D$ where each record contains an \texttt{opinions} field: a list of opinion tuples $(P, I, E)$ with $P \in \{\text{Positive},\text{Negative},\text{Neutral}\}$, $I \in \{\text{Strong},\text{Standard},\text{Average},\text{Weak}\}$, and $E$ the polar expression span (string or list of tokens).}
\KwOut{A filtered dataset $D'$ containing only reviews with a valid review-level polarity label, augmented with four new attributes: \texttt{Polar\_expression}, \texttt{Polarity}, \texttt{Intensity}, and \texttt{gold\_polarity\_label}.}

\ForEach{record $r \in D$}{
    $L \gets \text{parse}(r.\texttt{opinions})$ \tcp*{list of opinion dictionaries}
    \If{$L = \emptyset$}{
        set all new attributes to empty string \;
        \textbf{continue} \;
    }
    $r.\texttt{Polar\_expression} \gets \big[ \text{span}(E) \ \big|\ (P,I,E) \in L \big]$ \tcp*{use full span; join tokens if $E$ is a list}
    $r.\texttt{Polarity} \gets \text{join}\big( [P \mid (P,I,E) \in L], \text{`,`}\big)$ \;
    $r.\texttt{Intensity} \gets \text{join}\big( [I \mid (P,I,E) \in L], \text{`,`}\big)$ \;
    $\text{posScore} \gets 0,\ \text{negScore} \gets 0,\ \text{hasNeutral} \gets \text{false}$ \;
    \ForEach{$(P,I,E) \in L$}{
        \uIf{$P = \text{Positive}$}{
            \If{$I = \text{Strong}$}{$\text{posScore} \gets \text{posScore} + 2$}
            \Else{$\text{posScore} \gets \text{posScore} + 1$}
        }
        \uElseIf{$P = \text{Negative}$}{
            \If{$I = \text{Strong}$}{$\text{negScore} \gets \text{negScore} + 2$}
            \Else{$\text{negScore} \gets \text{negScore} + 1$}
        }
        \ElseIf{$P = \text{Neutral}$}{
            $\text{hasNeutral} \gets \text{true}$
        }
    }
    \uIf{$\text{posScore} > \text{negScore}$}{
        $r.\texttt{gold\_polarity\_label} \gets \text{``Positive''}$
    }
    \uElseIf{$\text{negScore} > \text{posScore}$}{
        $r.\texttt{gold\_polarity\_label} \gets \text{``Negative''}$
    }
    \ElseIf{$\text{posScore} = \text{negScore}$ \textbf{and} $(\text{posScore} > 0 \ \text{or} \ \text{hasNeutral})$}{
        $r.\texttt{gold\_polarity\_label} \gets \text{``Neutral''}$
    }
    \Else{
        $r.\texttt{gold\_polarity\_label} \gets \text{``''}$ \tcp*{no valid opinion}
    }
}
$D' \gets \{ r \in D : r.\texttt{gold\_polarity\_label} \neq \text{``''} \}$ \tcp*{remove discarded reviews}
\Return{$D'$ with review-level polarity labels}
\end{algorithm*}

\begin{table}[H]
\centering
\caption{Dataset statistics by split: total rows, empty labels, and remaining usable rows}
\label{tab:dataset_stats}
\small
\setlength{\tabcolsep}{4pt}
\renewcommand{\arraystretch}{0.95}
\begin{tabular}{l l c c c}
\toprule
\textbf{Dataset} & \textbf{Data} & \textbf{Total} & \textbf{Discarded} & \textbf{Remaining} \\
& \textbf{Split} & \textbf{Rows} & \textbf{Rows} & \textbf{Rows} \\
\midrule
\multirow{3}{*}{OpeNER\textsubscript{en}} & Train & 1744 & 344 & 1400 \\
           & Dev & 249  & 51  & 198  \\
           & Test & 499  & 92  & 407  \\
\midrule
\multirow{3}{*}{OpeNER\textsubscript{es}} & Train & 1438 & 186 & 1252 \\
           & Dev & 206  & 32  & 174  \\
           & Test & 410  & 48  & 362  \\
\midrule
\multirow{3}{*}{Darmstadt\_unis} & Train & 2253 & 1572 & 681 \\
                & Dev & 232  & 150  & 82  \\
                & Test & 318  & 214  & 104 \\
\midrule
\multirow{3}{*}{MPQA} & Train & 5619 & 4371 & 1248 \\
     & Dev & 1991 & 1575 & 416  \\
     & Test & 2042 & 1660 & 382  \\
\midrule
\multirow{3}{*}{MultiBooked\textsubscript{ca}} & Train & 1174 & 172 & 1002 \\
                & Dev & 167  & 27  & 140  \\
                & Test & 335  & 54  & 281  \\
\midrule
\multirow{3}{*}{MultiBooked\textsubscript{eu}} & Train & 1063 & 164 & 899 \\
                & Dev & 152  & 32  & 120  \\
                & Test & 305  & 65  & 240  \\
\midrule
\multirow{3}{*}{NoReC} & Train & 8634 & 4313 & 4321 \\
      & Dev & 1531 & 748  & 783  \\
      & Test & 1272 & 633  & 639  \\
\midrule
\multirow{3}{*}{MARC\textsubscript{en}} & Train & 200000 & 0 & 200000 \\
      & Dev & 5000 & 0  & 5000  \\
      & Test & 5000 & 0  & 5000  \\
\midrule
\multirow{3}{*}{MARC\textsubscript{es}} & Train & 200000 & 0 & 200000 \\
      & Dev & 5000 & 0  & 5000  \\
      & Test & 5000 & 0  & 5000  \\
\midrule
\multirow{3}{*}{MARC\textsubscript{fr}} & Train & 200000 & 0 & 200000 \\
      & Dev & 5000 & 0  & 5000  \\
      & Test & 5000 & 0  & 5000  \\
\midrule
\multirow{3}{*}{MARC\textsubscript{de}} & Train & 200000 & 0 & 200000 \\
      & Dev & 5000 & 0  & 5000  \\
      & Test & 5000 & 0  & 5000  \\
\bottomrule
\end{tabular}
\end{table}

\section{Methodology}
\label{sec:methodology}

\begin{figure*}[t]
  \includegraphics[width=1\linewidth, height=10cm, keepaspectratio]
  {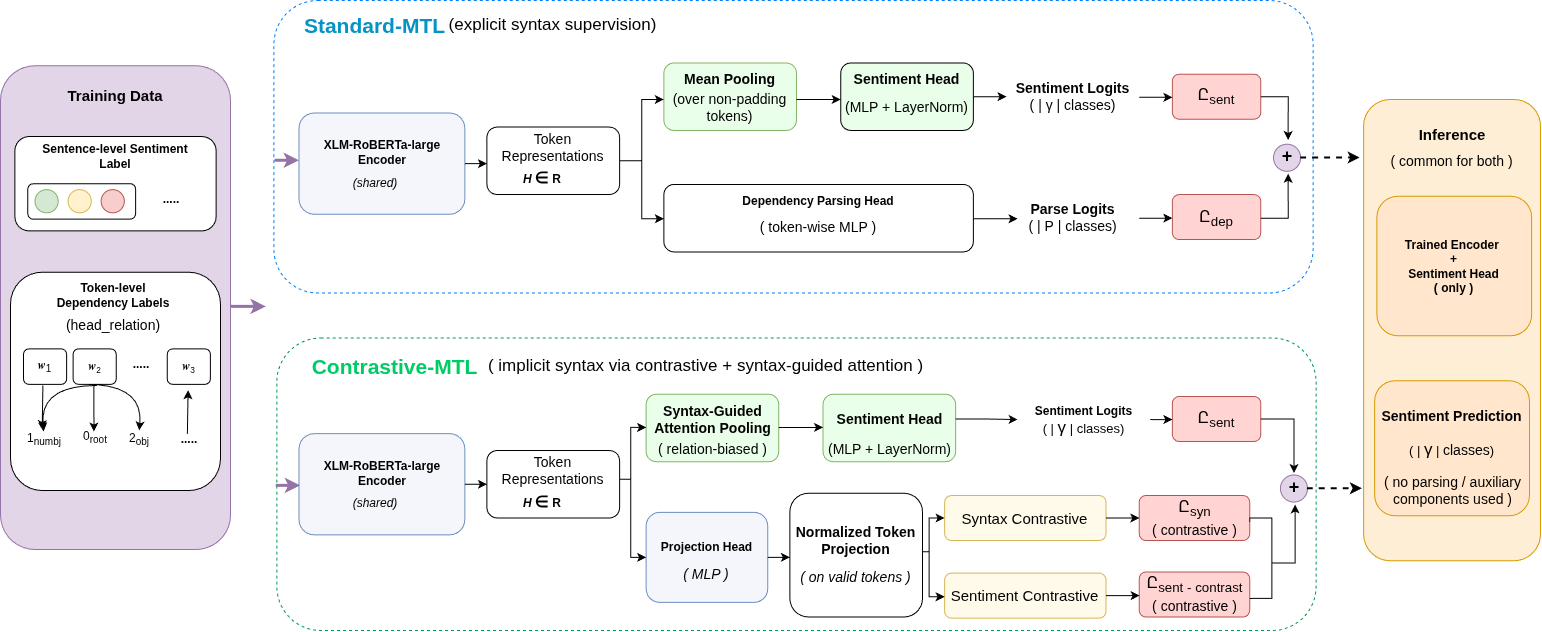}
  \caption{Overall architecture of the proposed Multi-Task Learning Framework: Joint Dependency Parsing (Standard-MTL) and Contrastive Syntax-Guided Attention (Contrastive-MTL)}
  \label{fig:methodology}
\end{figure*}

We propose two multi-task learning (MTL) strategies that integrate syntactic knowledge into a sentiment analysis model without requiring hand-crafted rules or external parsing at inference time. Both approaches share a common transformer encoder and a sentiment classification head, but differ in how they leverage automatically obtained dependency annotations during training as shown in Figure~\ref{fig:methodology}. The first, Standard-MTL, casts dependency parsing (sequence labeling) as an explicit auxiliary task, jointly trained with sentiment prediction. The second, Contrastive-MTL, integrates syntax implicitly through a contrastive auxiliary objective and a syntax-biased attention pooling mechanism. 
The auxiliary syntactic components—including the dependency parsing head in Standard-MTL, and the syntax-guided attention pooling together with the contrastive projection head in Contrastive-MTL—are discarded after training. At inference, both models use a standard sentiment head over a plain pooling mechanism; in particular, Contrastive-MTL replaces the syntax-guided attention with a learned attention pooling that has no access to dependency relation labels.

\subsection{Standard Multi-Task Learning with Syntactic Supervision}

The Standard-MTL model jointly learns to predict sentiment and dependency parses through explicit multi-task supervision. The architecture consists of a shared transformer encoder (XLM-RoBERTa-large, \cite{conneau2020unsupervised}) and two separate task heads. The transformer-based XLM-RoBERTa-large was selected as the backbone encoder due to its proven effectiveness in cross-lingual settings and its ability to produce rich contextualized token representations essential for both dependency parsing (sequence labeling) and sentiment classification.

Given an input batch, the encoder produces contextualized token representations $\mathbf{H} \in \mathbb{R}^{B \times L \times d}$, where $d$ is the hidden size. For sentiment classification, a sentence-level representation is obtained via mean pooling over all non-padding tokens:
\begin{equation*}
    \mathbf{h}_{\text{sent}} = \frac{\sum_{j=1}^L \mathbf{H}_{:,j,:} \cdot \mathbb{I}[M_{:,j}=1]}{\sum_{j=1}^L \mathbb{I}[M_{:,j}=1]}
\end{equation*}
where $M$ is the attention mask. The pooled vector is passed through a sentiment classification head $\text{Head}_{\text{sent}}$, which consists of LayerNorm, a two-layer MLP with GELU activation and dropout (0.2), and a final linear projection to $|\mathcal{Y}|$ classes. We used the same architecture in our baseline model as well for the fair comparison. 

For the auxiliary task, a dependency parsing head $\text{Head}_{\text{dep}}$ operates on every token's representation, consisting of a two-layer MLP: $\text{Linear}(d, d) \to \text{GELU} \to \text{Dropout}(0.1) \to \text{Linear}(d, |\mathcal{P}|)$.

The overall loss is a weighted sum of a label-smoothed cross-entropy loss for sentiment and a standard cross-entropy loss for dependency parsing, with $\lambda$ fixed to $0.3$, where invalid positions (sub-tokens, padding) are ignored:
\begin{equation*}
    \mathcal{L}_{\text{MTL}} = \mathcal{L}_{\text{sent}}( \hat{y}, y ) \;+\; \lambda \cdot \mathcal{L}_{\text{dep}}( \hat{\mathbf{p}}, \mathbf{p} )
\end{equation*}

The sentiment loss uses label smoothing with $\epsilon = 0.1$:
\begin{equation*}
    \mathcal{L}_{\text{sent}} = -\sum_{c=1}^{|\mathcal{Y}|} q_c \log \hat{p}_c
\end{equation*}
where $\hat{p} = \text{softmax}(\hat{y})$ is the predicted distribution, and $q_c = (1-\epsilon)\mathbb{I}[c=y] + \epsilon/|\mathcal{Y}|$ is the smoothed target distribution that mixes the one-hot true label with a uniform prior.

By jointly optimizing sentiment classification and dependency parsing, the shared encoder learns to capture syntactic regularities—such as negation scope and modifier attachments—that are crucial to determine the sentiment polarity. The auxiliary parsing task prevents the encoder from overfitting to superficial lexical cues, encouraging it to build representations that are more robust to word order and syntactic variation. Consequently, the sentiment head benefits from a syntax-enriched feature space, leading to improved performance without requiring any additional computation at inference time.

\subsection{Multi-Task Contrastive Learning with Syntax-Guided Attention}

The Contrastive-MTL variant integrates syntactic information implicitly through two mechanisms: a syntax-guided attention pooling module and a pair of contrastive auxiliary objectives operating on token-level representations. The sentiment classification backbone is identical to that of Standard-MTL, except for the pooling step.

\subsubsection{Syntax-Guided Attention Pooling}
Instead of mean pooling, we use a learnable attention mechanism that can be biased by the dependency relations of the tokens. Given token representations $\mathbf{H}$, attention scores are computed as:
\begin{equation*}
    \mathbf{a} = \text{softmax}\bigl( \mathbf{W}_a \mathbf{H} + \mathbf{b}^{\text{rel}}_{r_j} \bigr)
\end{equation*}
where $\mathbf{W}_a \in \mathbb{R}^{1 \times d}$ is a learnable linear layer, and $\mathbf{b}^{\text{rel}}_{r_j}$ is a scalar bias taken from an embedding lookup $\mathbf{E}_{\text{rel}} \in \mathbb{R}^{|\mathcal{R}| \times 1}$ indexed by the dependency relation label $r_j$ of token $j$. The relation-dependent bias $\mathbf{b}^{\text{rel}}_{r_j}$ is only available during training because it requires gold dependency relations. At inference, this term is omitted, and the model uses the learned linear attention $\mathbf{W}_a$ alone, i.e., $\mathbf{a} = \text{softmax}\bigl( \mathbf{W}_a \mathbf{H} \bigr)$.
The sentence representation is then the weighted sum:
\begin{equation*}
    \mathbf{h}_{\text{sent}} = \sum_{j=1}^L a_j \cdot \mathbf{H}_{:,j,:}
\end{equation*}
This allows the model to learn which syntactic roles are more informative for sentiment, guided by the relation labels. The pooled representation is then fed into the sentiment head ($\text{Head}_{\text{sent}}$). 

\subsubsection{Contrastive Auxiliary Objectives}
To further encourage the encoder to capture syntactic and sentiment regularities in the token space, we add contrastive losses on a projection head ($\text{Proj}$) that maps each token representation to a lower-dimensional space $\mathbb{R}^{d_{\text{proj}}}$ (with $d_{\text{proj}} = 256$):
\begin{equation*}
    \mathbf{Z} = \text{Proj}(\mathbf{H}) = \text{Linear}(\text{GELU}(\text{Linear}(\mathbf{H}, d)), d_{\text{proj}})
\end{equation*}
All token projections that correspond to valid first-subword positions are normalized to the unit hypersphere. Two contrastive losses are then computed:
\begin{enumerate}
    \item \textbf{Syntax contrastive loss} -- tokens sharing the same dependency relation label (extracted from the parse string) are treated as positive pairs; those with different labels are negatives.
    \item \textbf{Sentiment contrastive loss} -- tokens belonging to sentences with the same sentiment label are positives; others are negatives (token-level sentiment labels are simply the sentence label repeated for all valid tokens).
\end{enumerate}

For valid tokens with normalized projections $\mathbf{z}_i$ and class labels $c_i$, the contrastive loss computed as follows:

\begin{equation*}
    \mathcal{L}_{\text{contrast}} = -\frac{1}{|\mathcal{P}|}\sum_{i \in \mathcal{P}} \log \frac{\frac{1}{|\mathcal{P}_i|}\sum_{j \in \mathcal{P}_i} \exp(\mathbf{z}_i^{\top}\mathbf{z}_j / \tau)}{\sum_{k \neq i} \exp(\mathbf{z}_i^{\top}\mathbf{z}_k / \tau)}
\end{equation*}

where $\mathcal{P}_i = \{j \neq i : c_j = c_i\}$, $\mathcal{P} = \{i : \mathcal{P}_i \neq \emptyset\}$, and $\tau = 0.07$.
We compute this loss twice: once for $\mathcal{L}_{\text{syn}}$ (using dependency relation labels as the class) and once for $\mathcal{L}_{\text{sent-contrast}}$ (using sentence-level sentiment labels as the class). The positive term is normalized by \(|\mathcal{P}_i|\) per anchor \(i\), preventing many-positive anchors from dominating and stabilizing training under relation/sentiment class imbalance; this is applied identically to \(\mathcal{L}_{\text{syn}}\) and \(\mathcal{L}_{\text{sent-contrast}}\).

\subsubsection{Learnable Loss Weighting}
The two auxiliary contrastive losses are combined with the main sentiment loss using uncertainty-based weighting. Following~\cite{kendall2018multi}, we introduce two learnable scalar parameters $\sigma_{\text{syn}}$ and $\sigma_{\text{sent}}$, which control the relative contribution of the syntax and sentiment contrastive objectives. The total training loss is:
\begin{equation*}
    \mathcal{L}_{\text{total}} = \mathcal{L}_{\text{sent}} + \sum_{t \in \{\text{syn, sent-contrast}\}} \left(\frac{0.5}{\sigma_t^2} \mathcal{L}_t + \log \sigma_t\right)
\end{equation*}
At the start of training, $\sigma_{\text{syn}}$ and $\sigma_{\text{sent}}$ are initialized to $1$ (i.e., $\log \sigma = 0$). This formulation dynamically balances the magnitude of the auxiliary gradients, preventing one objective from dominating the optimization.

The contrastive auxiliary objectives and syntax-guided attention pooling implicitly distill syntactic knowledge into token representations. By pulling together tokens that share the same dependency relation or belong to sentences with the same sentiment, the model learns to associate specific syntactic roles (e.g., nsubj, amod) with sentiment expression, while the learnable attention bias further emphasizes relationally important tokens. This implicit syntactic shaping yields a more discriminative sentence representation, enabling the sentiment classifier to generalize better across syntactically diverse inputs—all without explicit parsing at the inference time.

\subsection{Training Procedure}

Table~\ref{tab:hyperparameters} summarizes the hyperparameters used to train the  models. We use the XLM-RoBERTa-large\footnote{\url{https://huggingface.co/FacebookAI/xlm-roberta-large}} encoder, initialized from pre-trained  weights, with task-specific heads randomly initialized. Training is performed  with the AdamW optimizer using two learning rates: $2\times 10^{-5}$ for the  encoder parameters and $1\times 10^{-4}$ for the head parameters. A linear  learning rate warmup is applied over the first 10\% of the total training steps, 
followed by linear decay. Gradient norms are clipped to a maximum of $1.0$.

We use a batch size of 16 and train for up to 50 epochs. Early stopping is  employed on the development set macro $F_1$ score with a patience of 7 epochs;  the model state with the best development $F_1$ is restored for final evaluation.  All experiments are run with a fixed random seed (42) for reproducibility.

\begin{table}[htbp]
\centering
\caption{Hyperparameters used to train the models}
\label{tab:hyperparameters}
\small
\begin{tabular}{l l}
\toprule
\textbf{Hyperparameter} & \textbf{Value} \\
\midrule
Encoder & XLM-RoBERTa-large \\
Encoder init. & Pre-trained \\
Head init. & Random \\
Optimizer & AdamW \\
Enc. learning rate & $2 \times 10^{-5}$ \\
Head learning rate & $1 \times 10^{-4}$ \\
LR schedule & Linear warmup + decay \\
Warmup & 10\% of steps \\
Gradient clipping & $1.0$ \\
Batch size & 16 \\
Max epochs & 50 \\
Early stopping & 7 epochs (dev $F_1$) \\
Random seed & 42 \\
\bottomrule
\end{tabular}
\end{table}

\subsection{Evaluation}

We measured the performance on the held-out test set and reported accuracy, macro-averaged precision, recall and $F_1$ scores. Since the auxiliary tasks are discarded at inference, the evaluation procedure is identical for all models: only the sentiment logits are computed, and no additional parsing or syntactic information is required.

\section{Results}
\subsection{Polarity Classification Performance}

\begin{figure*}[t]
  \centering
  \includegraphics[width=\linewidth]{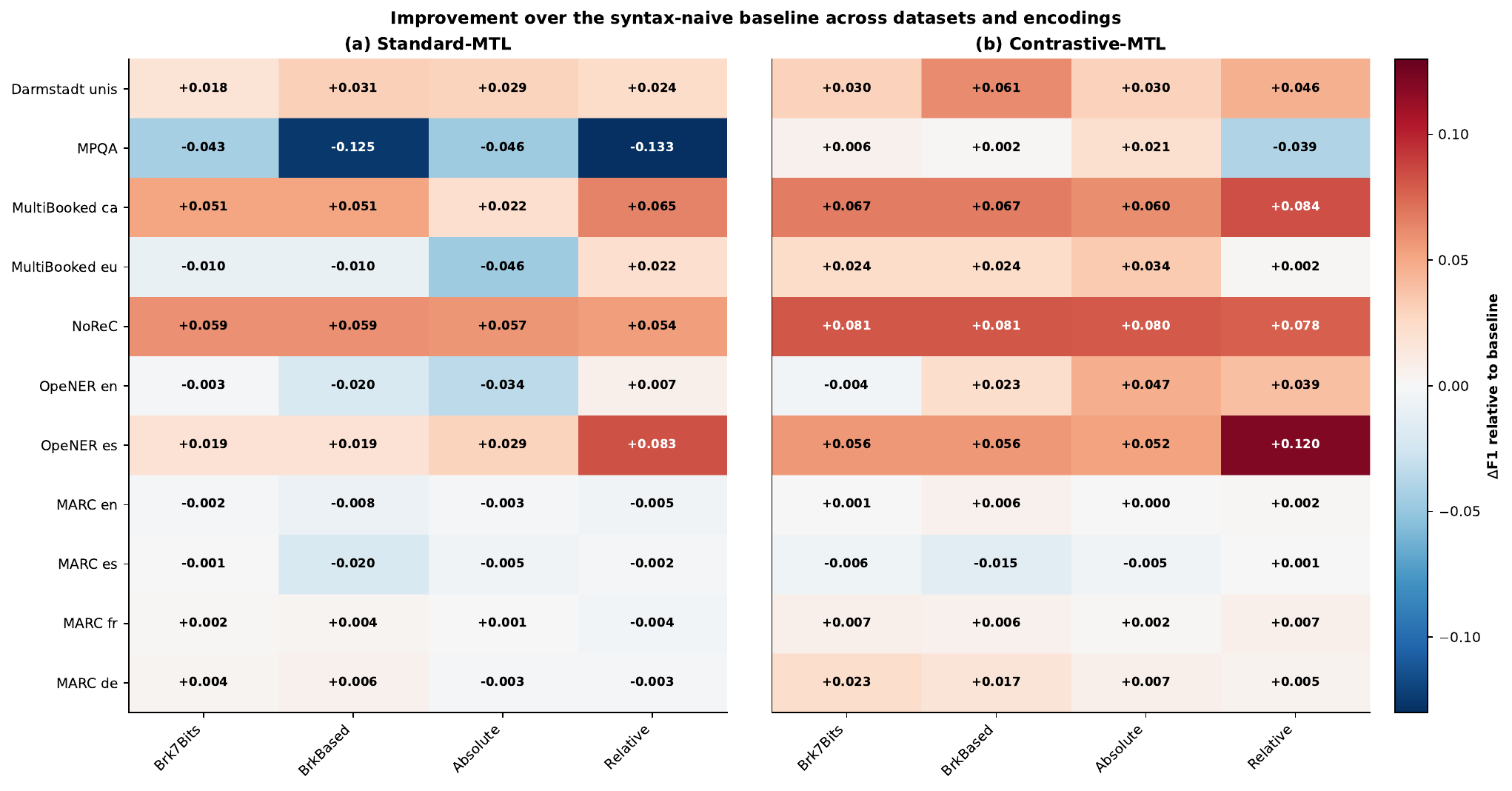}
  \caption{$\Delta$F1 relative to the syntax-naive baseline: (a) Standard-MTL and (b) Contrastive-MTL.}
  \label{fig:delta_heatmap}
\end{figure*}

\begin{figure*}[t]
  \centering
  \includegraphics[width=\linewidth]{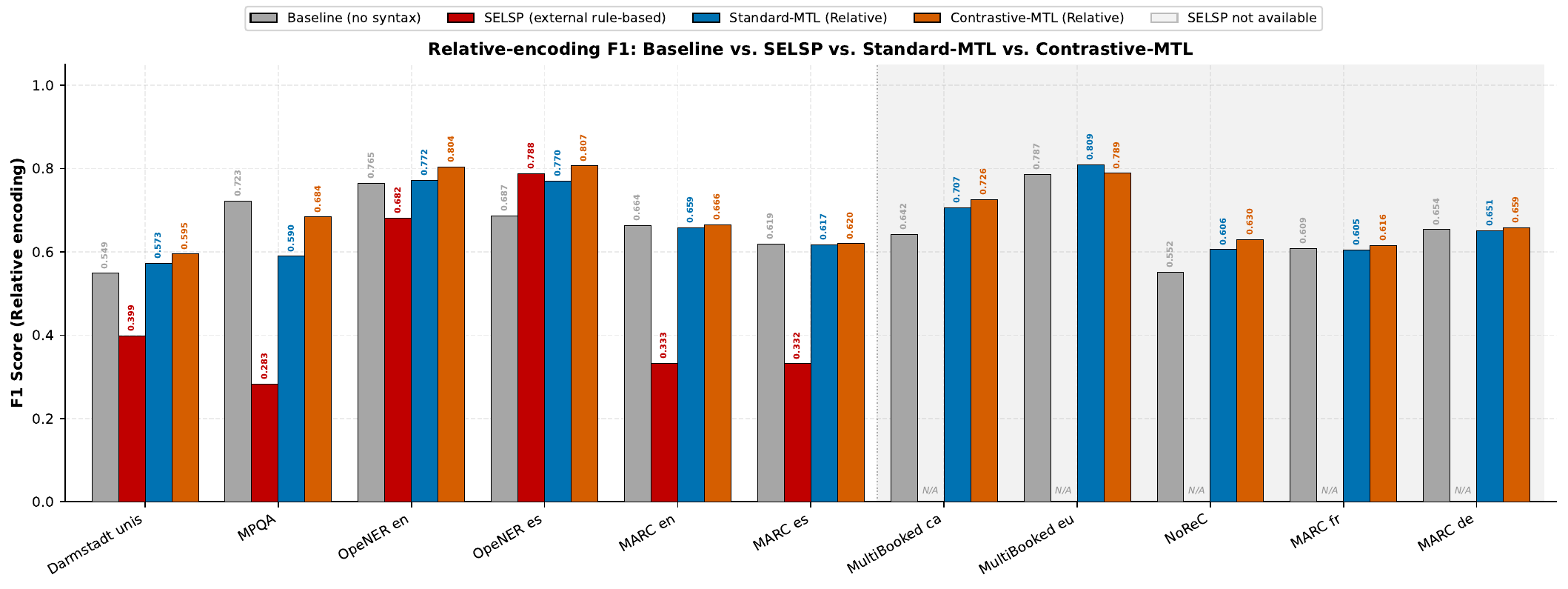}
  \caption{Relative-encoding F1 for Baseline, SELSP, Standard-MTL and Contrastive-MTL across all eleven datasets. SELSP is available only for the six English and Spanish datasets; the white region with N/A marks the five datasets for which SELSP is not applicable.}
  \label{fig:baseline_selsp_best}
\end{figure*}

\begin{figure*}[t]
  \centering
  \includegraphics[width=\linewidth]{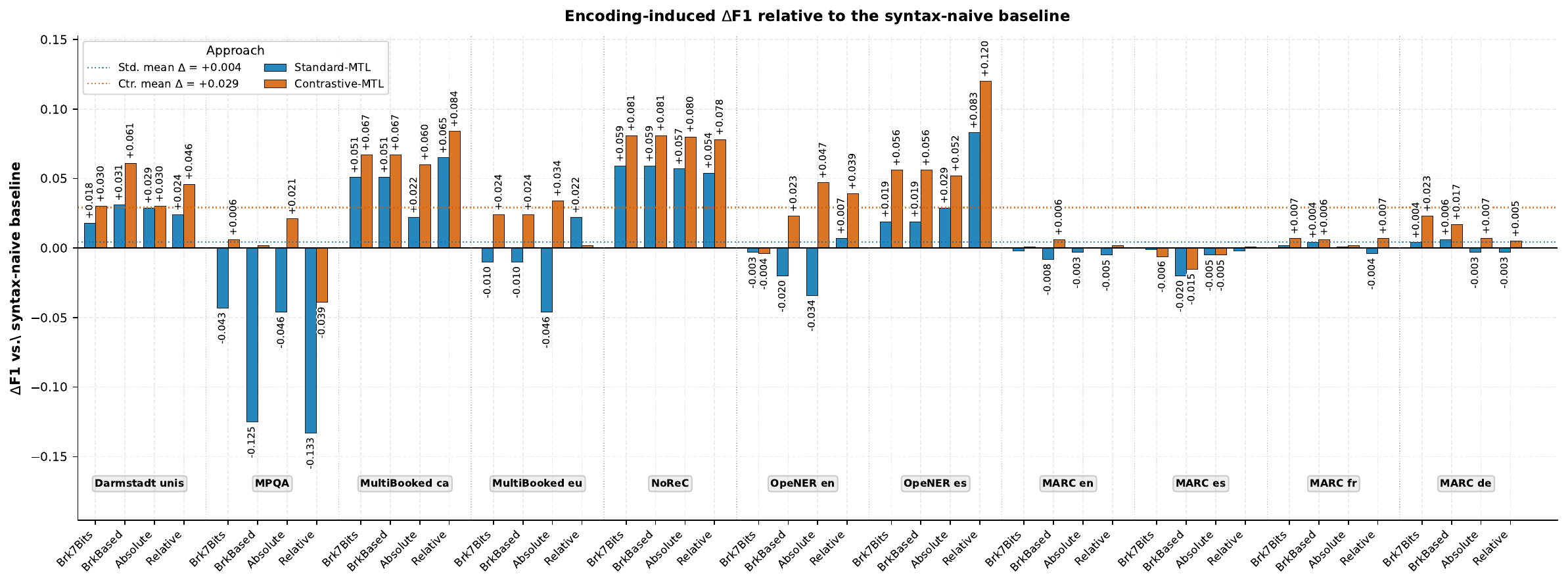}
  \caption{Encoding-induced $\Delta$F1 relative to the syntax-naive baseline, for Standard-MTL and Contrastive-MTL.}
  \label{fig:delta_from_baseline}
\end{figure*}

Table \ref{tab:performance-comparison} presents a comparison of Standard Multi-Task Learning (Standard-MTL) and Contrastive Multi-Task Learning (Contrastive-MTL), both incorporating syntactic knowledge, against a baseline model without syntactic information and an external rule-based pipeline (SELSP) that requires parsing at inference time. While we report results for all encoding schemes, our primary analysis focuses on the relative encoding, which achieves the best overall performance and provides the most consistent improvements over both the baseline and SELSP. Note that SELSP only supports English and Spanish; consequently, it could not be applied to the remaining five datasets, which are in Catalan, Basque, Norwegian, French, and German.

With the relative encoding, Contrastive-MTL outperforms the baseline on 10 out of 11 datasets. The only exception is MPQA, where the baseline remains ahead (0.723 vs. 0.684). This consistency contrasts with Standard-MTL, which beats the baseline on only 6 out of 11 datasets with the same encoding. The gap between the two MTL variants is most pronounced on datasets with rich syntactic structure: on MultiBooked\textsubscript{ca}, Contrastive-MTL improves over the baseline by +0.084 F1 versus +0.065 for Standard-MTL; on OpeNER\textsubscript{es}, the improvement is +0.120 versus +0.083. On the MARC datasets, differences are narrower but consistent, with Contrastive-MTL achieving the best F1-scores across all four language subsets. Figure~\ref{fig:delta_heatmap} visualises these $\Delta$F1 values as a heatmap for all four encodings. Contrastive-MTL is positive on at least 9 of 11 datasets under every encoding (10 of 11 under BrkBased and Relative), while Standard-MTL is positive on at most 6 of 11 under every encoding and shows clearly negative regions on MPQA and MultiBooked\textsubscript{eu}.

Turning to SELSP, which is available for the six English and Spanish datasets, Contrastive-MTL with the relative encoding outperforms it on all six, whereas Standard-MTL beats SELSP on five out of six, falling short only on OpeNER\textsubscript{es} (0.770 vs. 0.788) as visualized in Figure~\ref{fig:baseline_selsp_best}. This is particularly notable given that SELSP explicitly parses the input at inference time, while our MTL models discard the parser after training. On the two datasets where SELSP is most competitive---OpeNER\textsubscript{es} (SELSP F1 = 0.788) and OpeNER\textsubscript{en} (SELSP F1 = 0.682)---Contrastive-MTL with the relative encoding achieves F1-scores of 0.807 and 0.804, respectively, demonstrating that implicit syntactic supervision during training can match or exceed the performance of explicit parsing at inference.

Although the relative encoding yields the strongest overall results, the pattern is consistent across the four tested encodings. Contrastive-MTL beats the baseline on 9--10 out of 11 datasets depending on the encoding, while Standard-MTL beats it on only 5--6. Figure~\ref{fig:delta_from_baseline} shows this encoding-by-encoding robustness directly: Contrastive-MTL is at or above baseline on 39 of the 44 dataset--encoding combinations (with a single exact tie at MARC\textsubscript{en} under the Absolute encoding), and the only regressions are small ($|\Delta| \leq 0.04$, all on MPQA, OpeNER\textsubscript{en} and MARC\textsubscript{es}). In contrast, Standard-MTL drops below baseline in 21 of the 44 combinations, most severely on MPQA (down to $-0.133$). This indicates that the advantage of Contrastive-MTL is not an artifact of a particular encoding scheme but reflects a more effective integration of syntactic information.

\subsection{Computational Efficiency}

Table \ref{tab:time-comparison-averaged} reports training and evaluation times. SELSP is an external rule-based pipeline and is therefore not trained; its evaluation time is only reported for the English and Spanish datasets it supports. MTL variants require longer training times on some datasets relative to baseline, as seen on MultiBooked\textsubscript{eu} (652--664 vs.\ 263\,s) or OpeNER\textsubscript{es} (653--851 vs.\ 459\,s). However, the pattern reverses on other datasets: on NoReC, Standard-MTL trains faster than baseline (3026 vs.\ 4700\,s), and on OpeNER\textsubscript{en}, Contrastive-MTL is also faster (693 vs.\ 817\,s). Between MTL variants, Contrastive-MTL is often more expensive to train than Standard-MTL (e.g., MPQA and MARC\textsubscript{fr}), though not always (e.g., OpeNER\textsubscript{en} and Darmstadt\_unis). Inference times, by contrast, remain nearly identical across baseline, Standard-MTL, and Contrastive-MTL, since the parser is discarded after training.

\begin{figure*}[t]
  \centering
  \includegraphics[width=\linewidth]{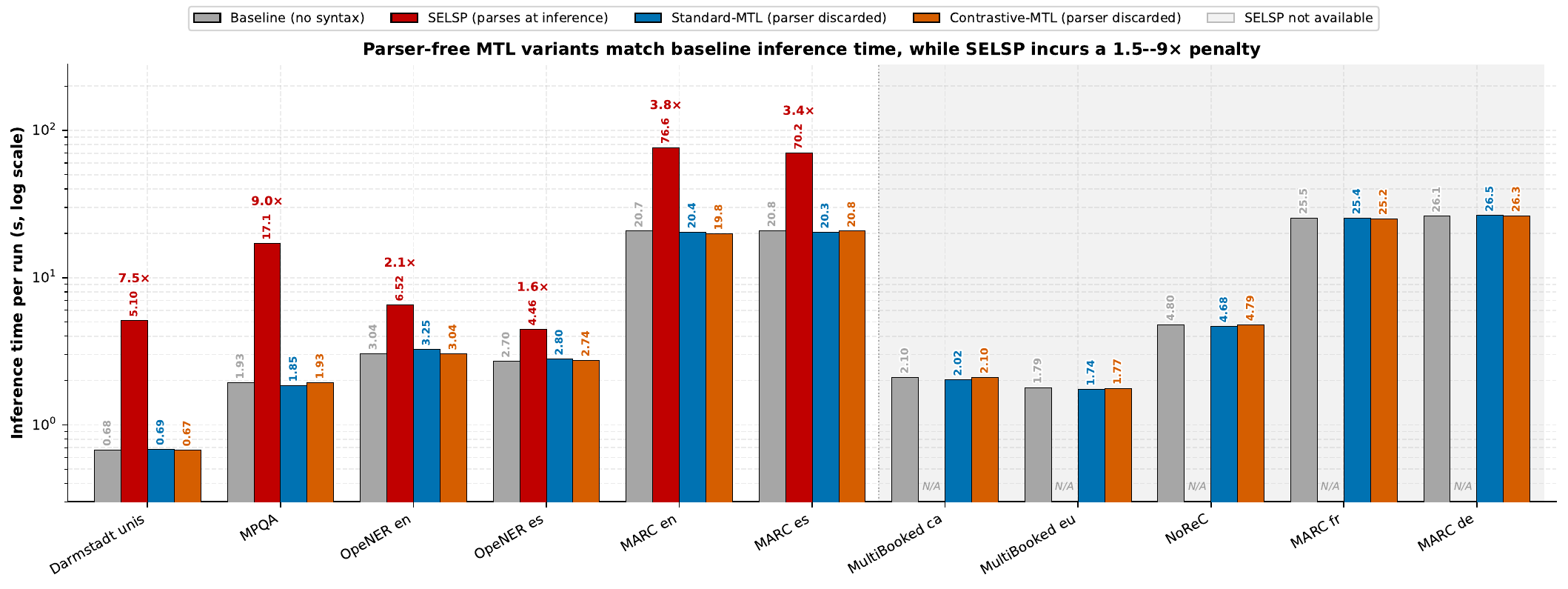}
  \caption{Inference time comparison (log scale) for Baseline, SELSP, Standard-MTL and Contrastive-MTL.}
  \label{fig:inference_time}
\end{figure*}

This is not the case for SELSP, which must parse the input at inference time. Figure~\ref{fig:inference_time} summarises the inference cost across all datasets: the parser free MTL variants are statistically indistinguishable from the baseline (mean deviation $< 6\%$), whereas SELSP is approximately $1.6$--$9.2\times$ slower depending on the dataset, with the penalty widest on MPQA ($9.2\times$) and Darmstadt\_unis ($7.4\times$) and still clearly visible on the larger MARC corpora ($\approx 3.7\times$). Thus, the MTL variants achieve competitive performance without the inference-time parsing overhead that makes SELSP costly.

\begin{table*}[!tp]
\centering
\caption{Performance comparison of explicit versus implicit syntactic augmentation. Standard Multi-Task Learning (Standard-MTL) and Contrastive Multi-Task Learning (Contrastive-MTL) are compared against a syntax-naive baseline and an external rule-based pipeline (SELSP) across all datasets and encoding schemes. SELSP results are only reported for the English and Spanish datasets, as the system does not support the remaining languages (Catalan, Basque, Norwegian, French, and German).}
\label{tab:performance-comparison}
\footnotesize
\setlength{\tabcolsep}{5pt}  
\renewcommand{\arraystretch}{0.85}
\begin{tabular}{l l 
  c c c c | c c c c}
\toprule
\textbf{Dataset} & \textbf{Encoding} & 
\multicolumn{4}{c}{\textbf{Standard-MTL}} & \multicolumn{4}{c}{\textbf{Contrastive-MTL}} \\
\cmidrule(lr){3-6} \cmidrule(lr){7-10}
& & \textbf{Acc} & \textbf{Prec} & \textbf{Rec} & \textbf{F1} & \textbf{Acc} & \textbf{Prec} & \textbf{Rec} & \textbf{F1} \\
\midrule
\multirow{5}{*}{Darmstadt\_unis} & Baseline & 0.822 & 0.546 & 0.553 & 0.549 & 0.822 & 0.546 & 0.553 & 0.549 \\
& SELSP & 0.414 & 0.581 & 0.414 & 0.399 & 0.414 & 0.581 & 0.414 & 0.399 \\
\cmidrule(lr){2-10}
& Brk7Bits & 0.843 & 0.574 & 0.562 & 0.567 & 0.865 & 0.575 & 0.582 & 0.579 \\
& BrkBased & 0.867 & 0.576 & 0.584 & 0.580 & 0.911 & 0.608 & 0.612 & \textbf{0.610} \\
& Absolute & 0.867 & 0.576 & 0.581 & 0.578 & 0.867 & 0.580 & 0.579 & 0.579 \\
& Relative & 0.856 & 0.570 & 0.582 & 0.573 & 0.889 & 0.591 & 0.601 & 0.595 \\
\midrule
\multirow{5}{*}{MPQA} & Baseline & 0.882 & 0.751 & 0.707 & 0.723 & 0.882 & 0.751 & 0.707 & 0.723 \\
& SELSP & 0.327 & 0.381 & 0.350 & 0.283 & 0.327 & 0.381 & 0.350 & 0.283 \\
\cmidrule(lr){2-10}
& Brk7Bits & 0.902 & 0.714 & 0.665 & 0.680 & 0.890 & 0.756 & 0.714 & 0.729 \\
& BrkBased & 0.891 & 0.590 & 0.606 & 0.598 & 0.887 & 0.745 & 0.707 & 0.725  \\
& Absolute & 0.898 & 0.710 & 0.662 & 0.677 & 0.891 & 0.810 & 0.716 & \textbf{0.744} \\
& Relative & 0.879 & 0.578 & 0.605 & 0.590 & 0.883 & 0.916 & 0.650 & 0.684 \\
\midrule
\multirow{5}{*}{MultiBooked\textsubscript{ca}} & Baseline & 0.886 & 0.643 & 0.641 & 0.642 & 0.886 & 0.643 & 0.641 & 0.642 \\
\cmidrule(lr){2-10}
& Brk7Bits & 0.904 & 0.707 & 0.683 & 0.693 & 0.904 & 0.704 & 0.716 & 0.709  \\
& BrkBased & 0.904 & 0.707 & 0.683 & 0.693 & 0.904 & 0.704 & 0.716 & 0.709  \\
& Absolute & 0.907 & 0.693 & 0.657 & 0.664 & 0.900 & 0.695 & 0.713 & 0.702 \\
& Relative & 0.889 & 0.695 & 0.739 & 0.707  & 0.907 & 0.715 & 0.748 & \textbf{0.726}\\
\midrule
\multirow{5}{*}{MultiBooked\textsubscript{eu}} & Baseline & 0.929 & 0.793 & 0.788 & 0.787 & 0.929 & 0.793 & 0.788 & 0.787 \\
\cmidrule(lr){2-10}
& Brk7Bits & 0.916 & 0.762 & 0.793 & 0.777 & 0.946 & 0.830 & 0.795 & 0.811 \\
& BrkBased & 0.916 & 0.762 & 0.793 & 0.777 & 0.946 & 0.830 & 0.795 & 0.811 \\
& Absolute & 0.929 & 0.801 & 0.709 & 0.741 & 0.946 & 0.865 & 0.786 & \textbf{0.821} \\
& Relative & 0.941 & 0.837 & 0.784 & 0.809 & 0.933 & 0.785 & 0.803 & 0.789 \\
\midrule
\multirow{5}{*}{NoReC} & Baseline & 0.835 & 0.550 & 0.555 & 0.552 & 0.835 & 0.550 & 0.555 & 0.552 \\
\cmidrule(lr){2-10}
& Brk7Bits & 0.817 & 0.607 & 0.626 & 0.611 & 0.842 & 0.652 & 0.624 & \textbf{0.633}\\
& BrkBased & 0.817 & 0.607 & 0.626 & 0.611 & 0.842 & 0.652 & 0.624 & \textbf{0.633} \\
& Absolute & 0.832 & 0.612 & 0.609 & 0.609 & 0.821 & 0.627 & 0.660 & 0.632 \\
& Relative & 0.812 & 0.613 & 0.628 & 0.606 & 0.826 & 0.623 & 0.639 & 0.630 \\
\midrule
\multirow{5}{*}{OpeNER\textsubscript{en}} & Baseline & 0.906 & 0.753 & 0.789 & 0.765 & 0.906 & 0.753 & 0.789 & 0.765 \\
& SELSP & 0.705 & 0.777 & 0.705 & 0.682 & 0.705 & 0.777 & 0.705 & 0.682 \\
\cmidrule(lr){2-10}
& Brk7Bits & 0.926 & 0.786 & 0.750 & 0.762 & 0.931 & 0.814 & 0.741 & 0.761 \\
& BrkBased & 0.909 & 0.753 & 0.738 & 0.745 & 0.926 & 0.791 & 0.784 & 0.788 \\
& Absolute & 0.912 & 0.749 & 0.717 & 0.731 & 0.926 & 0.791 & 0.844 & \textbf{0.812} \\
& Relative & 0.914 & 0.785 & 0.764 & 0.772 & 0.931 & 0.826 & 0.789 & 0.804 \\
\midrule
\multirow{5}{*}{OpeNER\textsubscript{es}} & Baseline & 0.925 & 0.717 & 0.666 & 0.687 & 0.925 & 0.717 & 0.666 & 0.687 \\
& SELSP & 0.790 & 0.814 & 0.790 & 0.788 & 0.790 & 0.814 & 0.790 & 0.788 \\
\cmidrule(lr){2-10}
& Brk7Bits & 0.923 & 0.716 & 0.698 & 0.706 & 0.928 & 0.770 & 0.722 & 0.743 \\
& BrkBased & 0.923 & 0.716 & 0.698 & 0.706 & 0.928 & 0.770 & 0.722 & 0.743 \\
& Absolute & 0.928 & 0.744 & 0.700 & 0.716 & 0.920 & 0.715 & 0.775 & 0.739 \\
& Relative & 0.939 & 0.794 & 0.750 & 0.770 & 0.936 & 0.808 & 0.821 & \textbf{0.807} \\
\midrule
\multirow{5}{*}{MARC\textsubscript{en}} & Baseline & 0.663 & 0.670 & 0.663 & 0.664 & 0.663 & 0.670 & 0.663 & 0.664 \\
& SELSP & 0.377 & 0.447 & 0.377 & 0.333 & 0.377 & 0.447 & 0.377 & 0.333 \\
\cmidrule(lr){2-10}
& Brk7Bits & 0.663 & 0.662 & 0.663 & 0.662 & 0.662 & 0.673 & 0.662 & 0.665 \\
& BrkBased & 0.662 & 0.655 & 0.662 & 0.656 & 0.669 & 0.670 & 0.669 & \textbf{0.670} \\
& Absolute & 0.660 & 0.664 & 0.660 & 0.661 & 0.665 & 0.662 & 0.665 & 0.664 \\
& Relative& 0.660 & 0.658 & 0.660 & 0.659 & 0.667 & 0.671 & 0.667 & 0.666 \\
\midrule
\multirow{5}{*}{MARC\textsubscript{es}} & Baseline & 0.619 & 0.619 & 0.619 & 0.619 & 0.619 & 0.619 & 0.619 & 0.619 \\
& SELSP & 0.384 & 0.450 & 0.384 & 0.332 & 0.384 & 0.450 & 0.384 & 0.332 \\
\cmidrule(lr){2-10}
& Brk7Bits & 0.620 & 0.618 & 0.620 & 0.618 & 0.611 & 0.614 & 0.611 & 0.613 \\
& BrkBased & 0.599 & 0.600 & 0.599 & 0.599 & 0.606 & 0.601 & 0.606 & 0.604 \\
& Absolute & 0.616 & 0.619 & 0.616 & 0.614 & 0.616 & 0.611 & 0.616 & 0.614 \\
& Relative & 0.617 & 0.621 & 0.617 & 0.617 & 0.616 & 0.623 & 0.616 & \textbf{0.620} \\
\midrule
\multirow{5}{*}{MARC\textsubscript{fr}} & Baseline & 0.608 & 0.611 & 0.608 & 0.609 & 0.608 & 0.611 & 0.608 & 0.609 \\
\cmidrule(lr){2-10}
& Brk7Bits & 0.611 & 0.617 & 0.611 & 0.611 & 0.611 & 0.621 & 0.611 & \textbf{0.616} \\
& BrkBased & 0.614 & 0.619 & 0.614 & 0.613 & 0.608 & 0.621 & 0.608 & 0.615 \\
& Absolute & 0.612 & 0.609 & 0.612 & 0.610 & 0.614 & 0.615 & 0.614 & 0.611 \\
& Relative & 0.612 & 0.606 & 0.612 & 0.605 & 0.617 & 0.615 & 0.617 & \textbf{0.616} \\
\midrule
\multirow{5}{*}{MARC\textsubscript{de}} & Baseline & 0.657 & 0.653 & 0.657 & 0.654 & 0.657 & 0.653 & 0.657 & 0.654 \\
\cmidrule(lr){2-10}
& Brk7Bits & 0.660 & 0.658 & 0.660 & 0.658 & 0.680 & 0.675 & 0.680 & \textbf{0.677} \\
& BrkBased & 0.665 & 0.662 & 0.665 & 0.660 & 0.670 & 0.673 & 0.670 & 0.671 \\
& Absolute & 0.652 & 0.651 & 0.652 & 0.651 & 0.665 & 0.660 & 0.665 & 0.661 \\
& Relative & 0.649 & 0.660 & 0.649 & 0.651 & 0.657 & 0.662 & 0.657 & 0.659 \\
\bottomrule
\end{tabular}
\end{table*}

\begin{table*}[!tp]
\centering
\caption{Ablation study isolating the contributions of Contrastive Learning and Syntax-Guided Attention: Performance of Contrastive-only MTL (contrastive loss only) Versus Syntax-only MTL (syntax bias only)}
\label{tab:ablation-comparison}
\footnotesize
\setlength{\tabcolsep}{5pt}  
\renewcommand{\arraystretch}{0.85}
\begin{tabular}{l l 
  c c c c | c c c c}
\toprule
\textbf{Dataset} & \textbf{Encoding} & 
\multicolumn{4}{c}{\textbf{Contrastive-only MTL}} & \multicolumn{4}{c}{\textbf{Syntax-only MTL}} \\
\cmidrule(lr){3-6} \cmidrule(lr){7-10}
& & \textbf{Acc} & \textbf{Prec} & \textbf{Rec} & \textbf{F1} & \textbf{Acc} & \textbf{Prec} & \textbf{Rec} & \textbf{F1} \\
\midrule
\multirow{5}{*}{Darmstadt\_unis} & Baseline & 0.822 & 0.546 & 0.553 & 0.549 & 0.822 & 0.546 & 0.553 & 0.549 \\
\cmidrule(lr){2-10}
& Brk7Bits   & 0.899 & 0.597 & 0.607 & \textbf{0.602} & 0.843 & 0.577 & 0.554 & 0.558 \\
& BrkBased   & 0.878 & 0.595 & 0.586 & 0.589 & 0.844 & 0.565 & 0.564 & 0.563 \\
& Absolute & 0.856 & 0.578 & 0.568 & 0.570 & 0.867 & 0.578 & 0.581 & 0.579 \\
& Relative & 0.876 & 0.592 & 0.581 & 0.586 & 0.867 & 0.578 & 0.581 & 0.579 \\
\midrule
\multirow{5}{*}{MPQA} & Baseline & 0.882 & 0.751 & 0.707 & 0.723 & 0.882 & 0.751 & 0.707 & 0.723 \\
\cmidrule(lr){2-10}
& Brk7Bits   & 0.875 & 0.580 & 0.596 & 0.587 & 0.875 & 0.576 & 0.596 & 0.585 \\
& BrkBased   & 0.895 & 0.731 & 0.708 & 0.718 & 0.898 & 0.709 & 0.663 & 0.677 \\
& Absolute & 0.875 & 0.711 & 0.766 & \textbf{0.737} & 0.902 & 0.738 & 0.717 & 0.726 \\
& Relative & 0.883 & 0.693 & 0.698 & 0.695 & 0.855 & 0.561 & 0.592 & 0.573 \\
\midrule
\multirow{5}{*}{MultiBooked\textsubscript{ca}} & Baseline & 0.886 & 0.643 & 0.641 & 0.642 & 0.886 & 0.643 & 0.641 & 0.642 \\
\cmidrule(lr){2-10}
& Brk7Bits   & 0.900 & 0.691 & 0.713 & 0.698 & 0.868 & 0.673 & 0.693 & 0.681 \\
& BrkBased   & 0.900 & 0.691 & 0.713 & 0.698 & 0.868 & 0.673 & 0.693 & 0.681 \\
& Absolute & 0.882 & 0.679 & 0.700 & 0.686 & 0.868 & 0.668 & 0.692 & 0.676 \\
& Relative & 0.914 & 0.733 & 0.703 & \textbf{0.717} & 0.911 & 0.710 & 0.720 & 0.714 \\
\midrule
\multirow{5}{*}{MultiBooked\textsubscript{eu}} & Baseline & 0.929 & 0.793 & 0.788 & 0.787 & 0.929 & 0.793 & 0.788 & 0.787 \\
\cmidrule(lr){2-10}
& Brk7Bits   & 0.921 & 0.780 & 0.734 & 0.755 & 0.925 & 0.799 & 0.818 & \textbf{0.808} \\
& BrkBased   & 0.921 & 0.780 & 0.734 & 0.755 & 0.892 & 0.811 & 0.796 & 0.803 \\
& Absolute & 0.941 & 0.821 & 0.797 & 0.802 & 0.933 & 0.794 & 0.790 & 0.786 \\
& Relative & 0.925 & 0.799 & 0.818 & \textbf{0.808} & 0.933 & 0.807 & 0.730 & 0.763 \\
\midrule
\multirow{5}{*}{NoReC} & Baseline & 0.835 & 0.550 & 0.555 & 0.552 & 0.835 & 0.550 & 0.555 & 0.552 \\
\cmidrule(lr){2-10}
& Brk7Bits   & 0.826 & 0.609 & 0.622 & 0.612 & 0.809 & 0.585 & 0.608 & 0.595 \\
& BrkBased   & 0.826 & 0.609 & 0.622 & 0.612 & 0.809 & 0.585 & 0.608 & 0.595 \\
& Absolute & 0.835 & 0.601 & 0.599 & 0.600 & 0.832 & 0.645 & 0.615 & 0.625 \\
& Relative & 0.837 & 0.615 & 0.603 & \textbf{0.630} & 0.835 & 0.626 & 0.590 & 0.604 \\
\midrule
\multirow{5}{*}{OpeNER\textsubscript{en}} & Baseline & 0.906 & 0.753 & 0.789 & 0.765 & 0.906 & 0.753 & 0.789 & 0.765 \\
\cmidrule(lr){2-10}
& Brk7Bits   & 0.919 & 0.786 & 0.777 & 0.781 & 0.931 & 0.854 & 0.783 & 0.805 \\
& BrkBased   & 0.914 & 0.769 & 0.764 & 0.765 & 0.914 & 0.786 & 0.806 & 0.794 \\
& Absolute & 0.924 & 0.795 & 0.796 & \textbf{0.795} & 0.916 & 0.813 & 0.774 & 0.788 \\
& Relative & 0.912 & 0.793 & 0.750 & 0.765 & 0.914 & 0.772 & 0.778 & 0.774 \\
\midrule
\multirow{5}{*}{OpeNER\textsubscript{es}} & Baseline & 0.925 & 0.717 & 0.666 & 0.687 & 0.925 & 0.717 & 0.666 & 0.687 \\
\cmidrule(lr){2-10}
& Brk7Bits   & 0.923 & 0.736 & 0.724 & 0.727 & 0.925 & 0.736 & 0.740 & 0.738 \\
& BrkBased   & 0.923 & 0.736 & 0.724 & 0.727 & 0.925 & 0.736 & 0.740 & 0.738 \\
& Absolute & 0.898 & 0.718 & 0.855 & 0.741 & 0.928 & 0.607 & 0.600 & 0.603 \\
& Relative & 0.939 & 0.792 & 0.723 & \textbf{0.753} & 0.920 & 0.808 & 0.682 & 0.720 \\
\midrule
\multirow{5}{*}{MARC\textsubscript{en}} & Baseline & 0.663 & 0.670 & 0.663 & 0.664 & 0.663 & 0.670 & 0.663 & 0.664 \\
\cmidrule(lr){2-10}
& Brk7Bits & 0.660 & 0.658 & 0.660 & 0.658 & 0.674 & 0.671 & 0.667 & \textbf{0.669} \\
& BrkBased & 0.662 & 0.663 & 0.662 & 0.658 & 0.666 & 0.663 & 0.666 & 0.663 \\
& Absolute & 0.664 & 0.663 & 0.664 & 0.661 & 0.665 & 0.666 & 0.665 & 0.665 \\
& Relative & 0.666 & 0.663 & 0.666 & 0.663 & 0.664 & 0.663 & 0.664 & 0.661 \\
\midrule
\multirow{5}{*}{MARC\textsubscript{es}} & Baseline & 0.619 & 0.619 & 0.619 & 0.619 & 0.619 & 0.619 & 0.619 & 0.619 \\
\cmidrule(lr){2-10}
& Brk7Bits & 0.606 & 0.624 & 0.606 & 0.608 & 0.608 & 0.610 & 0.608 & 0.608 \\
& BrkBased & 0.609 & 0.619 & 0.609 & 0.610 & 0.607 & 0.610 & 0.607 & 0.607 \\
& Absolute & 0.614 & 0.619 & 0.614 & \textbf{0.614} & 0.611 & 0.617 & 0.611 & 0.612 \\
& Relative & 0.604 & 0.609 & 0.604 & 0.605 & 0.616 & 0.612 & 0.616 & \textbf{0.614} \\
\midrule
\multirow{5}{*}{MARC\textsubscript{fr}} & Baseline & 0.608 & 0.611 & 0.608 & 0.609 & 0.608 & 0.611 & 0.608 & 0.609 \\
\cmidrule(lr){2-10}
& Brk7Bits & 0.614 & 0.622 & 0.614 & \textbf{0.616} & 0.614 & 0.618 & 0.614 & 0.614 \\
& BrkBased & 0.609 & 0.624 & 0.609 & 0.613 & 0.608 & 0.606 & 0.608 & 0.605 \\
& Absolute & 0.605 & 0.616 & 0.605 & 0.609 & 0.612 & 0.614 & 0.612 & 0.608 \\
& Relative & 0.604 & 0.620 & 0.604 & 0.608 & 0.603 & 0.607 & 0.603 & 0.602 \\
\midrule
\multirow{5}{*}{MARC\textsubscript{de}} & Baseline & 0.657 & 0.653 & 0.657 & 0.654 & 0.657 & 0.653 & 0.657 & 0.654 \\
\cmidrule(lr){2-10}
& Brk7Bits & 0.671 & 0.667 & 0.671 & 0.665 & 0.672 & 0.670 & 0.672 & 0.669 \\
& BrkBased & 0.668 & 0.668 & 0.668 & 0.664 & 0.672 & 0.672 & 0.672 & 0.668 \\
& Absolute & 0.675 & 0.674 & 0.675 & 0.671 & 0.660 & 0.658 & 0.660 & 0.657 \\
& Relative & 0.676 & 0.674 & 0.676 & \textbf{0.672} & 0.661 & 0.667 & 0.661 & 0.663 \\
\bottomrule
\end{tabular}
\end{table*}

\begin{table*}[!tp]
\centering
\caption{Computational cost comparison between explicit (Standard MTL) and implicit (Contrastive MTL) syntactic augmentation strategies, reporting training and evaluation times in seconds computed on NVIDIA RTX 6000 GPU. For each MTL variant, we report the mean over the four encodings (Brk7Bits, BrkBased, Absolute, Relative). SELSP is an external rule-based pipeline and is therefore not trained; its evaluation time is only reported for the English and Spanish datasets.}
\label{tab:time-comparison-averaged}
\small
\setlength{\tabcolsep}{5pt}
\renewcommand{\arraystretch}{0.85}
\begin{tabular}{l l 
  c c | c c}
\toprule
\textbf{Dataset} & \textbf{Method} & 
\multicolumn{2}{c}{\textbf{Standard-MTL}} & \multicolumn{2}{c}{\textbf{Contrastive-MTL}} \\
\cmidrule(lr){3-4} \cmidrule(lr){5-6}
& & \textbf{Train (s)} & \textbf{Eval (s)} & \textbf{Train (s)} & \textbf{Eval (s)} \\
\midrule

\multirow{2}{*}{Darmstadt\_unis} & Baseline & 206.061 & 0.678 & 206.061 & 0.678 \\
& SELSP & - & 5.104 & - & 5.104 \\
& Ours & 325.681 & 0.686 & 282.294 & 0.674 \\
\midrule
\multirow{2}{*}{MPQA} & Baseline & 486.676 & 1.930 & 486.676 & 1.930 \\
& SELSP & - & 17.056 & - & 17.056 \\
& Ours & 420.828 & 1.848 & 559.545 & 1.930 \\
\midrule
\multirow{2}{*}{MultiBooked\textsubscript{ca}} & Baseline & 516.558 & 2.100 & 516.558 & 2.100 \\
& Ours & 610.246 & 2.024 & 636.596 & 2.102 \\
\midrule
\multirow{2}{*}{MultiBooked\textsubscript{eu}} & Baseline & 263.059 & 1.795 & 263.059 & 1.795 \\
& Ours & 651.633 & 1.739 & 663.974 & 1.771 \\
\midrule
\multirow{2}{*}{NoReC} & Baseline & 4699.542 & 4.801 & 4699.542 & 4.801 \\
& Ours & 3026.135 & 4.680 & 3688.326 & 4.786 \\
\midrule
\multirow{2}{*}{OpeNER\textsubscript{en}} & Baseline & 816.542 & 3.045 & 816.542 & 3.045 \\
& SELSP & - & 6.519 & - & 6.519 \\
& Ours & 1005.175 & 3.254 & 693.360 & 3.038 \\
\midrule
\multirow{2}{*}{OpeNER\textsubscript{es}} & Baseline & 459.175 & 2.699 & 459.175 & 2.699 \\
& SELSP & - & 4.456 & - & 4.456 \\
& Ours & 653.489 & 2.797 & 850.568 & 2.744 \\
\midrule
\multirow{2}{*}{MARC\textsubscript{en}} & Baseline & 27756.875 & 20.705 & 27756.875 & 20.705 \\
& SELSP & - & 76.638 & - & 76.638 \\
& Ours & 26601.118 & 20.364 & 28436.375 & 19.840 \\
\midrule
\multirow{2}{*}{MARC\textsubscript{es}} & Baseline & 26643.287 & 20.829 & 26643.287 & 20.829 \\
& SELSP & - & 70.167 & - & 70.167 \\
& Ours & 26106.984 & 20.334 & 28490.370 & 20.754 \\
\midrule
\multirow{2}{*}{MARC\textsubscript{fr}} & Baseline & 17922.380 & 25.499 & 17922.380 & 25.499 \\
& Ours & 18737.200 & 25.378 & 24623.209 & 25.194 \\
\midrule
\multirow{2}{*}{MARC\textsubscript{de}} & Baseline & 23765.453 & 26.120 & 23765.453 & 26.120 \\
& Ours & 25088.874 & 26.456 & 24485.178 & 26.328 \\
\bottomrule
\end{tabular}
\end{table*}

\section{Ablation Study}
To isolate the contributions of the syntax bias and the contrastive losses, we define two ablation variants:
\begin{itemize}
    \item \textbf{Contrastive-only}: Removes the dependency-relation bias
    $\mathbf{b}^{\text{rel}}_{r_j}$ from the attention pooling, replacing the syntax-guided attention with plain learned attention
    $\mathbf{a} = \mathrm{softmax}(\mathbf{W}_a \mathbf{H})$. Both contrastive losses ($\mathcal{L}_{\text{syn}}$ and
    $\mathcal{L}_{\text{sent-contrast}}$) and the uncertainty-based loss
    weighting are retained. This isolates the contribution of the syntax-guided pooling.
    \item \textbf{Syntax-only}: Retains the
    syntax-guided attention pooling, but removes the projection head, both contrastive terms, and the learnable loss weights $\sigma_{\text{syn}}, \sigma_{\text{sent}}$, so the model is trained with $\mathcal{L}_{\text{sent}}$ only. This isolates the contribution of the contrastive objectives.
\end{itemize}


Table \ref{tab:ablation-comparison} presents the results of an ablation study comparing two configurations: a model with the contrastive objective removed ("Syntax-only MTL") and a model with the syntactic encoding removed ("Contrastive-only MTL"), evaluated against a shared baseline that uses neither component.


The Contrastive-only MTL variant exhibits the most consistent pattern: across most datasets and encodings, it matches or exceeds the shared baseline, and it frequently attains the best or tied-best F1. This trend is especially clear on Darmstadt\_unis, MultiBooked\_ca, NoReC, OpeNER\textsubscript{es}, and MARC\textsubscript{de}, where the contrastive-only model broadly improves over the baseline. On MPQA, by contrast, it underperforms the baseline on three of the four encodings (Brk7Bits, BrkBased, and Relative), recovering above it only with the Absolute encoding. On MultiBooked\textsubscript{eu} and MARC\textsubscript{es}, it still reaches best or tied-best scores in some settings, although these gains are less uniform. These results indicate that contrastive learning alone provides a robust and transferable benefit on many, though not all, datasets.

The Syntax-only MTL variant shows a more mixed pattern. It yields notable improvements on OpeNER\textsubscript{en} and MARC\textsubscript{en}, and matches the best contrastive-only result on MultiBooked\textsubscript{eu} and MARC\textsubscript{es} in selected encodings. However, it often degrades performance relative to the baseline, particularly on MARC\textsubscript{es}, MPQA, and MARC\textsubscript{fr}, as well as the Absolute encoding on OpeNER\textsubscript{es}. Thus, syntax-guided attention alone does not reliably improve performance.

A direct comparison between the two ablated modules shows that Contrastive-only MTL achieves the highest single-component scores in more settings than Syntax-only MTL (seven datasets versus two, with two tied), indicating that the contrastive objective is the more robust individual component. However, neither ablation reproduces the full model's performance: taking the best encoding per dataset, the full Contrastive-MTL model outperforms both ablations on ten of the eleven datasets. This indicates that the syntax bias, while not reliably beneficial in isolation, provides complementary gains when combined with the contrastive objectives.

\section{Mechanism Analysis}
\label{sec:diagnostic}

The ablation in Table~\ref{tab:ablation-comparison} establishes that the contrastive objective drives downstream performance, but it leaves open the mechanism by which it does so. We examine that mechanism on OpeNER\textsubscript{en} as a case study from three angles: (i) which syntactic roles the trained model attends to at inference time (\S\ref{sec:attention}); (ii) what syntactic signal the training-time attention bias learned and subsequently discarded (\S\ref{sec:bias}); and (iii) whether the encoder internalizes syntactic structure, as measured by a linear probe on frozen representations (\S\ref{sec:probing}).

\subsection{Attention Analysis}
\label{sec:attention}

\begin{figure*}[!tp]
\centering
\includegraphics[width=\linewidth]{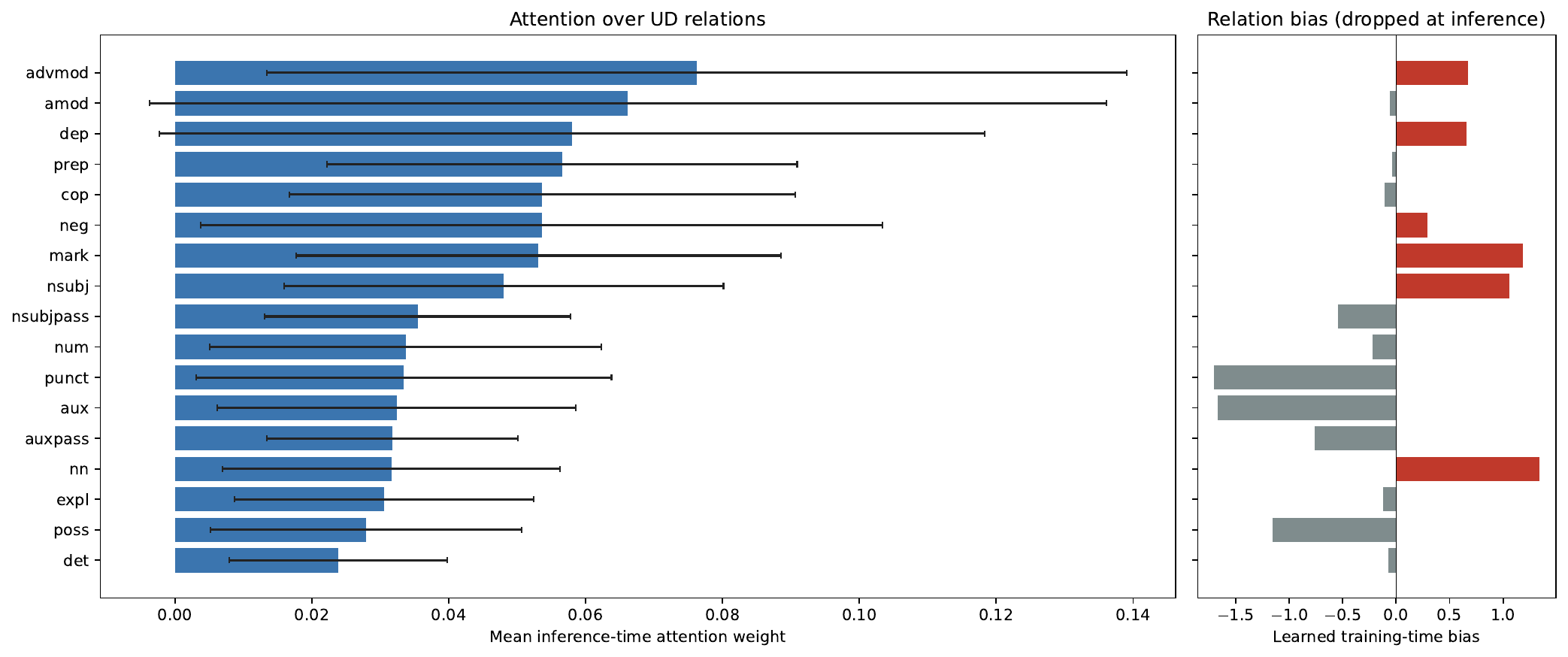}
\caption{Inference-time attention over UD relations on OpeNER\textsubscript{en} (Contrastive-MTL, Relative encoding, $n\geq 20$). Left panel: mean softmax attention weight per relation, with $\pm 1$ std across tokens; relations sorted by mean. Right panel: the learned
training-time relation bias (Section~\ref{sec:bias}), which is discarded at inference. Sentiment-relevant relations \emph{advmod}, \emph{amod}, and \emph{neg} receive high attention; function-word relations \emph{det} and \emph{poss} receive low attention.}
\label{fig:attention-opener}
\end{figure*}

Figure~\ref{fig:attention-opener} plots the mean inference-time attention weight assigned to each dependency relation, aggregated over all annotated test tokens as reported in Table~\ref{tab:attention-opener}. The highest attention is assigned to \emph{advmod} ($\bar{a} = 0.076$) and \emph{amod} ($\bar{a} = 0.066$)---the two relations that carry intensifiers (``very good'') and sentiment-bearing adjectives (``clean rooms''). The \emph{neg} ranks above the median ($\bar{a} = 0.054$) despite comprising only 63 test tokens. At the bottom of the distribution, the model assigns minimal attention to \emph{det} ($\bar{a} = 0.024$) and \emph{poss} ($\bar{a} = 0.028$)---the two function-word relations that carry no sentiment content. This is the pattern that rule-based sentiment systems predict: at inference time the model concentrates its pooling mass on the same syntactic roles that rule-based systems have hand-engineered for polarity composition. It is an emergent property of the encoder's learned representation under the contrastive objective---not of the syntax-guided attention bias, which is discarded at inference.

\subsection{Learned Relation Bias}
\label{sec:bias}

\begin{table}[!tp]
\centering
\caption{Inference-time attention and training-time learned bias per UD relation on OpeNER\textsubscript{en} (Contrastive-MTL, Relative encoding, $n\geq 20$).}
\label{tab:attention-opener}
\small
\begin{tabular}{l r c c}
\toprule
\textbf{Relation} & \textbf{$n$} & \textbf{Attention} & \textbf{Bias} \\
\midrule
\texttt{advmod}   & 320 & 0.0762 & $+0.673$ \\
\texttt{amod}     & 375 & 0.0662 & $-0.060$ \\
\texttt{dep}      &  40 & 0.0580 & $+0.660$ \\
\texttt{prep}     &  31 & 0.0565 & $-0.034$ \\
\texttt{cop}      & 265 & 0.0537 & $-0.107$ \\
\texttt{neg}      &  63 & 0.0536 & $+0.291$ \\
\texttt{mark}     &  61 & 0.0531 & $+1.187$ \\
\texttt{nsubj}    & 527 & 0.0480 & $+1.056$ \\
\texttt{nsubjpass}&  46 & 0.0354 & $-0.546$ \\
\texttt{num}      &  43 & 0.0337 & $-0.219$ \\
\texttt{punct}    &  76 & 0.0334 & $-1.701$ \\
\texttt{aux}      & 188 & 0.0324 & $-1.664$ \\
\texttt{auxpass}  &  50 & 0.0318 & $-0.764$ \\
\texttt{nn}       & 220 & 0.0316 & $+1.343$ \\
\texttt{expl}     &  21 & 0.0305 & $-0.120$ \\
\texttt{poss}     &  54 & 0.0279 & $-1.157$ \\
\texttt{det}      & 586 & 0.0238 & $-0.071$ \\
\bottomrule
\end{tabular}
\end{table}

The right panel of Figure~\ref{fig:attention-opener} shows the learned relation bias $\mathbf{b}^{\text{rel}}_{r}$ for each relation; the full per-relation values are reported in Table~\ref{tab:attention-opener}. Recall (Section~\ref{sec:methodology}) that this bias is present only during training and is discarded at inference; its magnitude and sign therefore show which relations the model \emph{wanted} to emphasize during training, providing a direct probe
of the syntactic knowledge the bias pathway absorbed.

The pattern is coherent with the attention analysis: \emph{advmod} receives the largest positive bias ($+0.67$), consistent with intensification being a reliable cue for sentiment polarity, and \emph{neg} receives a positive bias ($+0.29$), consistent with the model learning to upweight negation markers. \emph{amod} and \emph{det} both receive near-zero bias ($-0.06$ and $-0.07$), indicating that the model found these relations uninformative \emph{as a bias signal}---consistent with the fact that the polarity content of these relations is already carried by the token itself and does not need to be amplified at the pooling stage.

\subsection{Representation Probing}
\label{sec:probing}

To test whether the encoder itself internalizes syntax, we freeze the encoder and train a fresh linear probe on the frozen token representations to predict the gold UD relation of each annotated token. The probe is a single linear layer ($d{=}1024$ input, one output per relation class) trained for 30 epochs with AdamW (lr $10^{-3}$), batch size $1024$, and no regularization; it is trained on the training split and evaluated on the test split.

\begin{table*}[!tp]
\centering
\caption{Per-relation linear-probe accuracy on OpeNER\textsubscript{en} (Relative encoding. \emph{Lexical} relations are largely predictable from the token itself; \emph{context-dependent} relations require the surrounding sentence. Contrastive-MTL uses the attention bias and both contrastive losses; Contrastive-only MTL drops the bias; Syntax-only MTL drops the contrastive losses. Best per row in bold; relations with $n<20$ omitted.}
\label{tab:probe-per-relation}
\small
\begin{tabular}{l c r r r r}
\toprule
\textbf{Relation} & \textbf{$n$} & \textbf{Baseline} &
\multicolumn{1}{c}{\textbf{Contrastive-MTL}} &
\multicolumn{1}{c}{\textbf{Contrastive-only MTL}} &
\multicolumn{1}{c}{\textbf{Syntax-only MTL}} \\
\midrule
\multicolumn{6}{l}{\emph{Lexical relations (predictable from token identity)}} \\
det       & 586 & 0.804 & \textbf{0.988} & \textbf{0.988} & 0.930 \\
punct     &  76 & 0.790 & \textbf{0.974} & 0.961 & 0.263 \\
amod      & 375 & 0.627 & 0.880 & \textbf{0.899} & 0.307 \\
nn        & 220 & 0.218 & \textbf{0.832} & \textbf{0.832} & 0.077 \\
\midrule
\multicolumn{6}{l}{\emph{Context-dependent relations (require the sentence)}} \\
nsubj     & 527 & 0.685 & \textbf{0.941} & 0.920 & 0.368 \\
advmod    & 320 & 0.369 & 0.906 & \textbf{0.909} & 0.022 \\
cop       & 265 & 0.283 & 0.955 & \textbf{0.959} & 0.008 \\
aux       & 188 & 0.160 & 0.947 & \textbf{0.957} & 0.016 \\
neg       &  63 & 0.079 & \textbf{0.984} & \textbf{0.984} & 0.000 \\
mark      &  61 & 0.033 & 0.918 & \textbf{0.934} & 0.016 \\
auxpass   &  50 & 0.000 & 0.880 & \textbf{0.920} & 0.000 \\
nsubjpass &  46 & 0.000 & 0.174 & \textbf{0.783} & 0.000 \\
\midrule
\textbf{Overall} & 3{,}042 & 0.462 & 0.904 & \textbf{0.917} & 0.298 \\
\bottomrule
\end{tabular}
\end{table*}

Table~\ref{tab:probe-per-relation} reports per-relation accuracy for four models: Baseline (no auxiliary loss), Contrastive-MTL (attention bias + both contrastive losses), Contrastive-only MTL (both contrastive losses, no bias), and Syntax-only MTL (attention bias, no contrastive losses). Three findings stand out.

First, the baseline encoder is \emph{not} learning syntax. It scores near or above 0.79 on \emph{det} and \emph{punct}---relations that are predictable from the token itself (``the'', ``a'', ``.'')---and near chance on context-dependent relations (\emph{nsubj} 0.69, \emph{cop} 0.28, \emph{neg} 0.08, \emph{advmod} 0.37). The baseline's overall accuracy of 0.462 is therefore a lexical prior, not a syntactic signal.

Second, the contrastive objective creates syntactic structure. Both models trained with the contrastive losses---Contrastive-MTL and Contrastive-only MTL---raise accuracy on every context-dependent relation to above 0.90 (\emph{nsubj} 0.92--0.94, \emph{advmod} 0.91, \emph{cop} 0.96, \emph{neg} 0.98), a jump of between $+0.24$ and $+0.91$ absolute over the baseline. On relations that are trivially predictable from token identity (\emph{det}, \emph{punct}), the gain is small or zero, as expected. The contrastive objective specifically internalizes the syntactic distinctions that are not lexically derivable.

Third, the attention bias \emph{without} the contrastive objective degrades the \emph{encoder's} syntactic encoding, even though the ablation in Table~\ref{tab:ablation-comparison} shows that its downstream F1 remains competitive on OpeNER\textsubscript{en}. Syntax-only MTL achieves probe
accuracy of 0.298, well below the baseline's 0.462. The mechanism is a shortcut: with relation labels available to the pooling layer at each training step, the sentiment head can access relation information directly, and the encoder has no incentive to encode relations itself; sentiment fine-tuning then partially erases the pretrained syntactic features. The result is a model whose pooling layer knows which relations matter for sentiment but whose encoder knows less syntax than a model that was never trained on syntax at all. This is the same shortcut-learning phenomenon documented in the NLP literature \citep{gururangan-etal-2018-annotation, niven-kao-2019-probing}.

The three signals converge on a coherent account: Contrastive-MTL on OpeNER\textsubscript{en} produces an encoder that (i) attains high linear separability by dependency relation on the relations that matter for sentiment, (ii) prioritizes those same relations at inference time through attention, and (iii) treats function-word relations as low-attention, low-bias background. Taken together, these three findings support the mechanism proposed in Section~\ref{sec:methodology}: the contrastive objective shapes the encoder to encode the syntactic distinctions that matter for sentiment.

\section{Conclusion}

In this work, we explored two multi-task learning strategies for integrating syntactic 
information into neural polarity classification models without incurring the computational 
overhead of external parsers at inference time. Our proposed approaches---Standard-MTL, 
which jointly learns dependency parsing as sequence labeling alongside polarity 
classification, and Contrastive-MTL, which injects syntactic guidance implicitly through 
a contrastive auxiliary objective and syntax-biased attention pooling---were evaluated on 
eleven multilingual benchmark datasets across four encoding schemes.

With the relative encoding, Contrastive-MTL outperforms the syntactically-naive baseline 
on 10 out of 11 datasets and Standard-MTL on the same proportion, with the most pronounced 
gains on OpeNER\textsubscript{es} (+0.120 F1) and MultiBooked\textsubscript{ca} (+0.084 F1). 
Notably, with the same encoding, Contrastive-MTL also outperforms SELSP---an external 
rule-based pipeline that requires full parsing at inference time---on all six datasets 
where SELSP is applicable.

A mechanism analysis on OpeNER\textsubscript{en} clarifies the source of these gains. Linear probes on frozen encoder representations show that the contrastive objective raises probe accuracy on dependency relations from 0.46 (baseline) to 0.92, with the gain concentrated on context-dependent relations (\textit{nsubj}, \textit{advmod}, \textit{cop}, \textit{neg}) rather than on relations predictable from token identity alone (\textit{det}, \textit{punct}). Attention analysis shows that inference-time pooling mass on the Contrastive-MTL model concentrates on precisely these relations and minimizes function-word relations, while the learned training-time bias on \textit{advmod} and \textit{neg} is positive and on \textit{det} is near zero. These complementary signals support the interpretation that the contrastive objective shapes the encoder to encode the syntactic distinctions that matter for sentiment, and the sentiment head at inference reweights tokens by those distinctions.


Finally, both MTL variants discard the auxiliary syntactic components after training, so 
inference times remain nearly identical to the baseline, whereas SELSP is \(1.6\)--\(9.2\times\) 
slower at evaluation due to parsing overhead. This design preserves the efficiency of single-task models while offering a principled alternative to pipeline architectures that rely on full parse tree generation and hand-crafted compositional rules.

\section{Limitations and Future Directions}
Despite the promising results, we acknowledge the limitations of our study. First, the MTL models 
rely on syntactic annotations produced by an off-the-shelf biaffine dependency parser 
\cite{dozat2017deep} during training, so any systematic errors or biases in 
the parser's output may propagate to the MTL framework; the contrastive objective could, 
in principle, learn to cluster tokens around incorrect relation labels, potentially 
harming generalization.


These limitations open several avenues for future research. Extending the framework to 
fine-grained and aspect-based sentiment analysis is a natural next step, where token-level 
contrastive objectives aligned with aspect terms and opinion targets could enable the 
model to learn richer associations between UD relations (e.g., \textit{nsubj}, \textit{obj}, 
\textit{amod}) and specific sentiment expressions, thereby handling multi-aspect reviews 
with contrasting polarities. A second direction is to explore alternative syntactic 
representations---constituency parses, phrase-structure trees, or lightweight syntactic 
features such as negation scope markers and intensifier chains---under the same 
contrastive framework, in order to identify the optimal trade-off between linguistic 
richness and learning efficiency.


\section{Acknowledgments}
We acknowledge the European Research Council (ERC), which has funded this research under the Horizon Europe research and innovation programme (SALSA, grant agreement No 101100615), 
GAP (PID2022-139308OA-I00) funded by MICIU/AEI/10.13039/501100011033/ and by ERDF, EU; LATCHING (PID2023-147129OB-C21) funded by MICIU/AEI/10.13039/501100011033 and ERDF (EU), Ministry for Digital Transformation and Civil Service and “NextGenerationEU” PRTR under grant TSI-100925-2023-1, Xunta de Galicia (ED431C 2024/02),  and Centro de Investigación de Galicia ‘‘CITIC’’, funded by the Xunta de Galicia through the collaboration agreement between the Consellería de Cultura, Educación, Formación Profesional e Universidades and the Galician universities for the reinforcement of the research centres of the Galician University System (CIGUS). Furthermore, this research was supported by the International, Interdisciplinary and Intersectoral Information and Communications Technology PhD programme (3-i ICT) granted to CITIC and supported by the European Union through the Horizon 2020 research and innovation programme under a Marie Skłodowska-Curie agreement (H2020-MSCA-COFUND), GA 101034261.

\clearpage

\bibliography{custom}




\end{document}